\documentclass{article}
\usepackage{iclr2027_conference,times}
\usepackage{amsmath,amssymb}
\usepackage{booktabs,graphicx,array}
\usepackage[hidelinks]{hyperref}
\usepackage{url}
\usepackage{microtype}
\usepackage{xspace}
\usepackage{float}
\usepackage{placeins}

\renewcommand{\arraystretch}{1.03}

\newcommand{\TopK}{\operatorname{TopK}}
\newcommand{\Corr}{\operatorname{Corr}}

\title{Rethinking Cross-Channel Importance in Time-Series Forecasting}
\author{%
Yong-Hoon Choi \quad Kwang-Hyun Park \quad Youngjin Cho\\
Division of Robotics, Kwangwoon University\\
Seoul 01897, Republic of Korea\\
\texttt{yhchoi@kw.ac.kr \quad akaii@kw.ac.kr \quad yjaycho@kw.ac.kr}
}

\begin{document}
\iclrfinalcopy
\maketitle

\begin{abstract}
Cross-channel modeling is central to multivariate time-series forecasting, yet channels that are statistically related, predictively useful, and actually used by a trained forecaster are often treated as if they defined the same notion of importance. We show that they need not coincide. Cross-channel dependency structures change substantially across future offsets, and horizon-adaptive source selection improves a controlled Ridge predictor in 21 of 32 dataset--prediction-length conditions, with a mean gain of $5.16\%$. This selected-set signal also transfers to a matched nonlinear predictor. Yet imposing the same horizon-specific source logic on iTransformer yields only 11 of 20 wins and a mean gain of $0.208\%$, with little alignment between controlled and neural gains. Functional interventions further show that strong forecasters use cross-channel information, while their source-reliance rankings agree little with controlled utility or with one another across iTransformer, TimesNet, and a cross-channel TimeMixer. As a constructive consequence, bounded post-hoc support improves a frozen channel-independent forecaster in 12 of 16 dataset--horizon conditions, with a positive aggregate bootstrap interval. Cross-channel importance should therefore be interpreted relative to the forecasting mechanism and question that define it: \textbf{related $\neq$ useful $\neq$ used}.
\end{abstract}

\section{Introduction}
Multivariate time-series forecasting relies on information distributed across variables. Modern models therefore devote substantial capacity to cross-channel interaction through attention, mixing, routing, clustering, or learned dependency structure. This raises a basic question: \emph{which channels matter for predicting a target variable?}

That question hides three different objects. A channel may be statistically related to the target, it may improve prediction under a specified forecasting mechanism, or a trained forecaster may actually rely on it. These are often treated as if they formed one chain,
\[
\text{dependency}\;\longrightarrow\;\text{predictive utility}\;\longrightarrow\;\text{model reliance},
\]
but neither implication is guaranteed. A related source can be redundant given the target history; a source that helps one predictor may add little to another; and different forecasters may organize the same multivariate information differently. This paper tests those two links rather than assuming them.

We first test whether observed cross-channel structure is stable across the forecast interval and find broad cross-offset variation. We then ask whether horizon-specific source utility transfers across forecasting mechanisms. Ridge-derived adaptive source sets remain useful under a matched nonlinear MLP, but imposing the same source logic as a sparse iTransformer topology provides little consistent benefit. This contrast motivates our central distinction between observed dependency $A$, controlled predictive utility $P$, and functional reliance $I^{\mathcal F}$.

Direct interventions confirm that strong neural forecasters use cross-channel information, yet their functional source rankings agree little with controlled utility or with one another. A same-checkpoint analysis further shows that this mismatch cannot be explained solely by independently trained prediction-length checkpoints. We finally test a constructive alternative to imposing an external source ranking as neural topology: using cross-channel information only as a bounded post-hoc correction to a frozen channel-independent forecast. Together, these tests ask whether cross-channel importance is a single dataset property or instead depends on the question and forecasting mechanism used to define it.

\begin{figure}[!t]
\centering
\includegraphics[width=\linewidth]{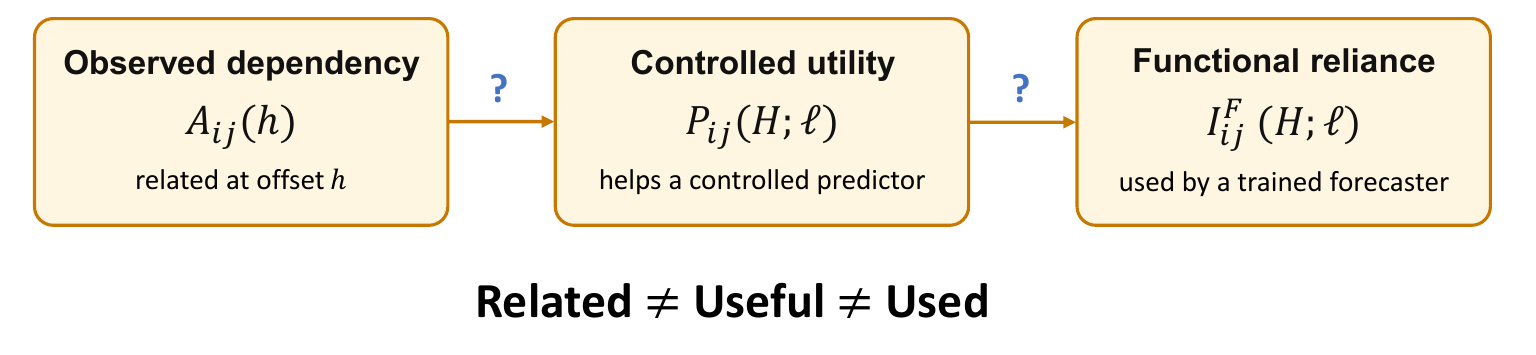}
\caption{\textbf{Three distinct notions of cross-channel importance.} The question marks denote links that must be tested rather than assumed: being related need not imply being predictively useful, and being useful need not imply being functionally used by a trained forecaster.}
\label{fig:overview}
\end{figure}

Figure~\ref{fig:overview} summarizes the resulting view. Our contributions are: \textbf{(1)} we establish broad cross-offset variation in observed dependency; \textbf{(2)} we show that Ridge-derived horizon-specific source sets can remain useful under a matched nonlinear controlled predictor without reliably transferring to a strong neural topology; \textbf{(3)} we empirically disentangle dependency, controlled utility, and functional reliance, including a same-checkpoint future-offset test and high-dimensional protocol-sensitivity controls; and \textbf{(4)} we show, with matched interventions across three neural architectures, that functional source reliance is forecaster-dependent, and provide a constructive study of bounded support that preserves the original forecast as a fallback.

\section{Related Work}
\paragraph{Forecasting backbones and channel modeling.}
Long-horizon forecasting has progressed through efficient Transformer and decomposition models such as Informer, Autoformer, and FEDformer \citep{zhou2021informer,wu2021autoformer,zhou2022fedformer}, as well as simpler or alternative backbones including DLinear, PatchTST, TimesNet, and TimeMixer \citep{zeng2023dlinear,nie2023patchtst,wu2023timesnet,wang2024timemixer}. These works also expose different treatments of the channel dimension. PatchTST is explicitly channel-independent, whereas Crossformer and iTransformer model cross-dimension or variate-level interactions \citep{zhang2023crossformer,liu2024itransformer}; TSMixer additionally studies hybrid channel mixing \citep{ekambaram2023tsmixer}. Our work does not propose another generic backbone; it studies what an inferred cross-channel structure actually represents.

\paragraph{Adaptive cross-channel structure.}
Recent methods increasingly avoid a single static all-to-all dependency. LIFT estimates local lead--lag relations and selectively exploits leading indicators \citep{zhao2024lift}; SOFTS aggregates channels through a global series core \citep{han2024softs}; and DUET combines temporal clustering with sparsified channel soft clustering \citep{qiu2025duet}. TimeFilter performs patch-specific graph filtration \citep{hu2025timefilter}, FACT models fine-grained across-variable interactions in time and frequency domains \citep{wang2026fact}, xCPD performs input-aware channel-patch routing \citep{li2026xcpd}, and TiWeaver targets fine-grained asynchronous inter-channel dependencies \citep{li2026tiweaver}. QDF separately shows that future prediction steps can constitute heterogeneous forecasting tasks \citep{wang2026qdf}. We therefore do not claim that dynamic, local, or horizon-sensitive structure is itself novel. Our question is whether observed dependency, predictive source utility, and the structure functionally used by a trained forecaster are interchangeable.

\paragraph{Feature importance and intervention.}
Permutation-based importance measures model reliance by disrupting an input and measuring the resulting loss change, with roots in random-forest variable importance \citep{breiman2001randomforests}. Dependence among inputs complicates naive permutation interpretations \citep{strobl2008conditional}, and model-reliance work explicitly notes that different well-performing models can assign different importance to the same variable \citep{fisher2019mcr}. Related interpretability studies also caution against equating internal attention patterns with explanations, while subsequent work emphasizes that such claims depend on the definition and diagnostic protocol \citep{jain2019attention,wiegreffe2019attention}. Our source-history intervention preserves each donor channel's temporal trajectory, averages over common donor mappings, and is interpreted only as \emph{functional reliance of a fixed trained forecaster}, not as causal influence in the data-generating process.

\section{Three Notions of Cross-Channel Importance}
Let $\mathbf X_t\in\mathbb R^{L\times C}$ be a history ending at forecast origin $t$, and let the prediction length be $H$. We distinguish $H$ from a future offset $h\in\{1,\ldots,H\}$. For target $i$ and source $j\neq i$, we study three quantities.

\paragraph{Observed dependency.}
We use training-only cross-lag correlation,
\begin{equation}
A_{ij}(h)=\left|\Corr\!\left(x_j(t),x_i(t+h)\right)\right|.
\label{eq:A}
\end{equation}
For sparse analysis, sources are ranked by $A_{ij}(h)$ and self-connections are excluded. This answers: \emph{which channels appear related to the target at offset $h$?}

\paragraph{Controlled predictive utility.}
Let $f_i$ be a target-history baseline and $f_{i\leftarrow j}$ the same controlled predictor augmented with source $j$. Under loss $\ell$,
\begin{equation}
P_{ij}(H;\ell)=100\,\frac{L_i(f_i)-L_i(f_{i\leftarrow j})}{L_i(f_i)}.
\label{eq:P}
\end{equation}
The primary source-selection experiment uses a regularized endpoint predictor with compact target/source history summaries; source features are residualized against target-history features before estimating incremental benefit. For source budget $K_{\rm sel}$,
\begin{equation}
\begin{aligned}
\mathcal S_{i,P}(H)&=\TopK_{j\neq i}P_{ij}(H;\ell_{\rm end}),\\
\mathcal S_{i,P}^{\rm shared}&=\TopK_{j\neq i}\;\mathbb E_H[P_{ij}(H;\ell_{\rm end})].
\end{aligned}
\label{eq:sets}
\end{equation}
This answers: \emph{which source improves a specified predictor under a specified objective?}

\paragraph{Functional reliance.}
For evaluation sample $n$, source $j$ is replaced by the same channel from donor sample $\pi(n)$,
\begin{equation}
\widetilde{\mathbf X}_{j,\pi}^{(n)}[:,k]=
\begin{cases}
\mathbf X^{(\pi(n))}[:,j], & k=j,\\
\mathbf X^{(n)}[:,k], & k\neq j.
\end{cases}
\label{eq:intervene}
\end{equation}
This preserves the source trajectory's internal temporal structure while breaking its alignment with the current target context. We define Expected Permutation Importance (EPI) as
\begin{equation}
\bar I_{ij}^{\mathcal F}(H;\ell)=\frac{1}{|\Pi|}\sum_{\pi\in\Pi}100\,\frac{L_i(\mathcal F(\widetilde{\mathbf X}_{j,\pi}))-L_i(\mathcal F(\mathbf X))}{L_i(\mathcal F(\mathbf X))}.
\label{eq:I}
\end{equation}
Full-horizon MSE is the primary reliance metric, with full-horizon MAE as robustness. EPI answers: \emph{which source information does this trained forecaster actually depend on?} Complete feature construction and intervention details are in Appendix~\ref{app:details}.

\paragraph{Evaluation protocol.}
Table~\ref{tab:protocol} summarizes the benchmark dimensions and the two sparse budgets. Dependency-drift analysis uses a broader neighborhood to detect structural change, whereas controlled selection and strong-model intervention require a genuinely sparse subset. For the general datasets we use $K_{\rm drift}=\min(20,\max(2,\lceil0.1(C-1)\rceil))$; PeMS uses $K_{\rm drift}=20$. Selection uses $K_{\rm sel}=\min(10,\lceil(C-1)/2\rceil)$, yielding $K_{\rm sel}=3$ for ETT rather than selecting all six non-target channels. All dependency estimation and source selection use training data only.

\begin{table}[t]
\centering
\small
\caption{\textbf{Datasets and sparse budgets.} $K_{\rm drift}$ is used only for cross-offset drift; $K_{\rm sel}$ is used for controlled selection and strong-model intervention.}
\label{tab:protocol}
\begin{tabular}{lrrrr}
\toprule
Dataset & $C$ & $H$ & $K_{\rm drift}$ & $K_{\rm sel}$\\
\midrule
PeMS03 & 358 & 12/24/48 & 20 & 10\\
PeMS04 & 307 & 12/24/48 & 20 & 10\\
PeMS07 & 883 & 12/24/48 & 20 & 10\\
PeMS08 & 170 & 12/24/48 & 20 & 10\\
Electricity & 321 & 96/192/336/720 & 20 & 10\\
Weather & 21 & 96/192/336/720 & 2 & 10\\
Solar & 137 & 96/192/336/720 & 14 & 10\\
ETTh1 & 7 & 96/192/336/720 & 2 & 3\\
ETTm1 & 7 & 96/192/336/720 & 2 & 3\\
\bottomrule
\end{tabular}
\end{table}

\section{Cross-Channel Structure Changes Across Future Offsets}
For each target, we compare sparse dependency neighborhoods across representative offsets $\{1,\mathrm{round}(H/4),\mathrm{round}(H/2),H\}$. Because low overlap can also reflect estimation noise, we recompute neighborhoods on two temporally interleaved training subsets and subtract same-offset instability. With Jaccard similarity $J(\cdot,\cdot)$, our primary adjusted drift is
\begin{equation}
D_{i,K}^{\rm adj}(h_a,h_b)=1-J(\mathcal N_{i,K}(h_a),\mathcal N_{i,K}(h_b))-\tfrac12\!\left[N_{i,K}(h_a)+N_{i,K}(h_b)\right],
\label{eq:drift}
\end{equation}
where $N_{i,K}(h)$ is the split-to-split same-offset instability.

\begin{figure}[H]
\centering
\includegraphics[width=.94\linewidth]{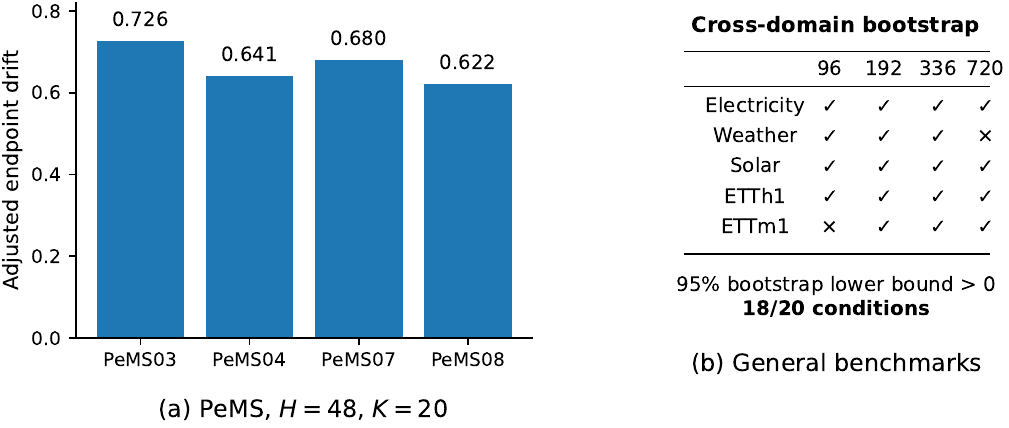}
\caption{\textbf{Cross-offset dependency drift.} PeMS exhibits large noise-adjusted endpoint drift at $H=48$ (left); for PeMS04, Top-10 neighborhoods at offsets 1 and 48 are nearly disjoint ($J=0.0031$). Across the general benchmarks, the 95\% bootstrap lower bound is positive in 18/20 conditions (right).}
\label{fig:drift}
\end{figure}

Figure~\ref{fig:drift} summarizes the result. At $H=48$ with primary $K=20$, adjusted endpoint drift is $0.726$, $0.641$, $0.680$, and $0.622$ on PeMS03/04/07/08. Under $K=10$, PeMS04 shows almost complete endpoint replacement, $J(1,48)=0.0031$. Beyond traffic, 18 of 20 general dataset--prediction-length conditions have a strictly positive 95\% bootstrap lower bound for adjusted endpoint drift. The magnitude is not universally monotonic in $H$; the phenomenon is broad but domain dependent. This matters for interpretation: cross-offset drift is evidence that the relation measured by Eq.~\ref{eq:A} changes with the forecasting task, not evidence that every longer horizon must have a progressively different graph. In particular, Weather and Solar include non-monotonic conditions, while ETT and PeMS show several pronounced endpoint shifts. We therefore use drift only as the motivation for the predictive test below, rather than treating it as a forecasting objective by itself.

\section{Is Horizon-Specific Structure Predictively Useful?}
\textbf{Yes---under controlled predictors, but not universally.} Observed drift is descriptive, so we next ask whether adapting the selected sources actually reduces forecasting error. Using the same Ridge predictor, source budget, train/test origins, and endpoint-MSE objective, Adaptive selection wins in 21 of 32 dataset--prediction-length conditions, with mean pooled gain $5.157\%$ and median gain $1.036\%$. The gains can be substantial: ETTh1 improves by $15.12\%$ and $17.26\%$ at $H=336$ and $720$, and ETTm1 improves by $9.45\%$ at $H=720$. Thus, the horizon-specific ranking is not merely a descriptive by-product of the dependency drift in Section~4.

\textbf{The selected-set signal is not limited to a linear final predictor.} We keep the Ridge-derived Shared and Adaptive source sets fixed and change only the final controlled forecaster to a small MLP with two 128-unit hidden layers. On the 20 general forecasting conditions shared with the strong-model study, Adaptive wins 13/20, with mean gain $2.137\%$ and median gain $1.852\%$. More importantly, condition-wise Ridge and MLP gains are strongly rank-aligned ($\rho=0.713$; Figure~\ref{fig:transfer}). Across all 32 conditions and three matched MLP seeds, Adaptive wins 19/32 with mean gain $1.667\%$ (Appendix~\ref{app:controlled}). Because the source sets remain defined by Ridge utility, this is a nonlinear transfer confirmation rather than an independent MLP definition of $P$; it weakens the simple explanation that the utility--topology gap is only a linear-versus-neural predictor mismatch.

The low-dimensional correction remains essential. With fixed $K=10$, ETTh1 and ETTm1 would select all six non-target variables, making Shared and Adaptive identical by construction. The dimension-aware $K_{\rm sel}$ in Table~\ref{tab:protocol} restores a genuine sparse comparison. In high-dimensional Electricity and PeMS04, additional controls also show that the magnitude---and in some conditions the sign---of the pooled Adaptive gain can vary with the evaluated 32-target subset and the source-candidate cap. We therefore interpret $P$ and its aggregate gains as \emph{protocol-conditioned} rather than as a unique dataset-level source graph; the original predeclared protocol remains the primary comparison (Appendix~\ref{app:highdim-robustness}).

\section{Does Horizon-Specific Utility Transfer to a Strong Forecaster?}
\textbf{Not reliably.} We next test whether the same training-only source rankings define a better sparse topology for iTransformer. DenseFineTune, SharedSparse, and HorizonAdaptiveSparse start from the same Phase-1 checkpoint. Only the final encoder block and forecast projection are fine-tuned with matched optimization. The sparse variants use the same $K_{\rm sel}$ and self-edge; HorizonAdaptiveSparse uses the ranking for the current prediction length, whereas SharedSparse uses a horizon-invariant Borda consensus of the four training-only rankings (Appendix~\ref{app:hardmask}). Let $\mathcal S_i^{\rm mask}(H)$ denote the source set supplied by either sparse policy. In the final encoder block,
\begin{equation}
M_{ij}(H)=\begin{cases}0,&j=i\ \text{or}\ j\in\mathcal S_i^{\rm mask}(H),\\-\infty,&\text{otherwise}.\end{cases}
\label{eq:mask}
\end{equation}
The mask therefore blocks direct attention from excluded sources in the intervened layer, while earlier frozen blocks remain unchanged.

We expanded the original diagnostic screen to \textbf{all 20 dataset--prediction-length conditions for which the same Phase-1 iTransformer checkpoint suite is available}. Under full-horizon MSE, HorizonAdaptiveSparse beats SharedSparse in 11/20 conditions, but the mean advantage is only $+0.208\%$ (median $+0.083\%$). More importantly, the controlled Ridge gain does not predict the neural gain: their condition-wise Spearman correlation is only $\rho=0.057$, with matching gain signs in 10/20 conditions. Matching the controlled endpoint objective does not restore the relation: endpoint MSE also gives 11/20 wins, mean gain $-0.120\%$, and $\rho=-0.074$ (Appendix~\ref{app:hardmask}).

\begin{figure}[!t]
\centering
\includegraphics[width=.98\linewidth]{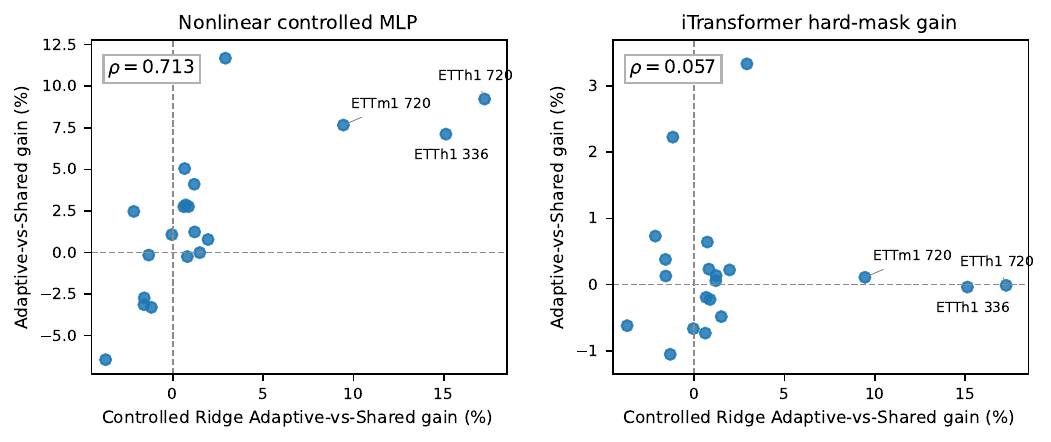}
\caption{\textbf{Predictive utility transfers across controlled predictors, but not reliably into iTransformer topology.} Each panel uses the same 20 general forecasting conditions. Left: Adaptive-vs-Shared gains under the matched nonlinear MLP are strongly aligned with the controlled Ridge gains ($\rho=0.713$). Right: the corresponding iTransformer hard-mask gains are nearly uncorrelated with Ridge ($\rho=0.057$).}
\label{fig:transfer}
\end{figure}

The contrast with the nonlinear controlled experiment is direct in Figure~\ref{fig:transfer}. Across the same 20 conditions, Ridge gains align strongly with the matched MLP ($\rho=0.713$) but are almost unrelated to the iTransformer hard-mask gains ($\rho=0.057$). The mismatch is especially visible on ETTh1: controlled gains of $15.123\%$ and $17.264\%$ at $H=336$ and $720$ become $-0.038\%$ and $-0.013\%$ under the iTransformer topology intervention. ETTm1 at $H=720$ similarly falls from $9.449\%$ controlled gain to $0.109\%$ neural gain.

Four softer matched mechanisms also fail to show a consistent positive transfer signal (Appendix~\ref{app:alternative}). The hard mask is retained in the main paper because it provides the cleanest topology-level test. \textbf{Takeaway: predictive utility can persist across controlled predictors without reliably transferring to the source topology of a strong forecaster.}

\section{What Does a Forecaster Actually Use?}
\subsection{Strong Models Use Cross-Channel Information}
A failed transfer would be uninformative if iTransformer simply ignored other channels. It does not. Replacing all non-target histories with histories from a common donor increases full-horizon MSE in all eight grouped-perturbation conditions, with mean degradation $15.842\%$. The pattern can be distributed: on Electricity at $H=192$, all-other perturbation increases endpoint MSE by $53.315\%$ while the mean evaluated single-source effect is only $0.029\%$. On Solar at $H=720$, the corresponding values are $19.412\%$ and $8.783\%$. Thus, strong cross-channel reliance need not be represented by a small list of individually critical variables.

\subsection{Do Dependency or Utility Identify the Sources a Model Uses?}
Across 737 candidate target--source pairs and 55 target--condition instances in eight representative conditions, Table~\ref{tab:individual-align} compares endpoint functional reliance with controlled utility, raw-space dependency, and final-layer iTransformer latent dependency. No static pairwise score recovers the full functional ranking: median target-wise Spearman correlations are $-0.125$, $0.022$, and $0.029$. Raw and latent dependency show somewhat better Top-5 overlap than $P$, so the result is not that relation measures contain no information; rather, none should be interpreted as the forecaster's functional source ranking.

\begin{table}[t]
\centering
\small
\caption{\textbf{Individual-source alignment with endpoint functional importance.} Spearman is the median over 55 target--condition instances; Jaccard@5, Recall@5, and NDCG@5 are means.}
\label{tab:individual-align}
\begin{tabular}{lrrrr}
\toprule
Signal & Spearman & Jaccard@5 & Recall@5 & NDCG@5\\
\midrule
Controlled utility $P$ & -0.125 & 0.350 & 0.447 & 0.436\\
Raw dependency $A_{\rm raw}$ & 0.022 & 0.417 & 0.531 & 0.576\\
Latent dependency $A_{\rm latent}$ & 0.029 & 0.448 & 0.560 & 0.560\\
\bottomrule
\end{tabular}
\end{table}

A direct same-checkpoint test removes a key ambiguity in cross-prediction-length comparisons. We fix one $H=720$ iTransformer checkpoint per dataset and evaluate source reliance at future offsets $h\in\{96,192,336,720\}$ using the same target--source pools, test windows, and perturbation mappings. Table~\ref{tab:same-checkpoint-main} shows that cross-offset EPI rankings are only moderately stable, while offset-specific predictive utility and EPI remain negatively aligned at all four offsets. Thus, the utility--reliance mismatch is visible even when model parameters are fixed; it is not explained solely by independently trained prediction-length checkpoints.

\begin{table}[t]
\centering
\small
\caption{\textbf{Same-checkpoint $H=720$ test.} Left: stability of iTransformer EPI source rankings across future offsets. Right: offset-specific alignment between predictive utility $P$ and functional reliance $I$.}
\label{tab:same-checkpoint-main}
\begin{minipage}[t]{0.54\linewidth}
\centering
\begin{tabular}{lrr}
\toprule
Offsets & Median $\rho(I_a,I_b)$ & Jaccard@5\\
\midrule
96--192  & 0.482 & 0.483\\
96--336  & 0.148 & 0.463\\
96--720  & 0.392 & 0.551\\
192--336 & 0.366 & 0.509\\
192--720 & 0.163 & 0.463\\
336--720 & 0.364 & 0.487\\
\bottomrule
\end{tabular}
\end{minipage}\hfill
\begin{minipage}[t]{0.40\linewidth}
\centering
\begin{tabular}{lrr}
\toprule
Offset & Median $\rho(P,I)$ & Jaccard@5\\
\midrule
96  & -0.115 & 0.339\\
192 & -0.276 & 0.355\\
336 & -0.121 & 0.401\\
720 & -0.235 & 0.326\\
\bottomrule
\end{tabular}
\end{minipage}
\end{table}

\subsection{The Utility--Reliance Mismatch Is Robust}
With 16 common donor mappings, objective-matched full-horizon utility, and an expanded random-source pool, the alignment remains weak or negative (Appendix~\ref{app:utility-robustness}). In the Extended pool, median $P$--iTransformer correlation is $-0.354$ under full-horizon MSE and $-0.320$ under full-horizon MAE. Thus, objective mismatch and candidate-pool restriction do not recover positive rank alignment.

\subsection{Functional Reliance Is Forecaster-Dependent}
Most importantly, ``used'' is not forecaster-invariant in the evaluated models. We add a third neural architecture, TimeMixer with explicit cross-channel mixing (\texttt{channel\_independence=0}), and reuse exactly the iTransformer/TimesNet target--source pairs, 512 test origins, and 16 donor mappings on Electricity $H=192$, Solar $H=720$, and ETTh1 $H=720$. Grouped all-other perturbation increases full-MSE for all 19 evaluated TimeMixer targets, confirming that the model functionally uses cross-channel information. Its donor/window split-half reliabilities are $0.921/0.888$, $0.951/0.696$, and $0.886/0.943$ on the three conditions, respectively.

Despite this reproducibility, the three neural forecasters do not recover a common source-reliance ranking. Median target-wise full-MSE EPI correlations are
\[
\rho(I^{\rm iT},I^{\rm TN})=-0.091,\qquad
\rho(I^{\rm iT},I^{\rm TM})=-0.009,\qquad
\rho(I^{\rm TN},I^{\rm TM})=0.103.
\]
Controlled utility is likewise weakly aligned with TimeMixer reliance, $\rho(P,I^{\rm TM})=0.053$. These values use identical source pairs and intervention mappings across the neural models; full reliability and training details are in Appendices~\ref{app:epi-reliability} and~\ref{app:timesnet}.

\textbf{Takeaway: cross-channel information is used, but which source is ``important'' is neither recovered by $A$ or $P$ nor invariant across the evaluated forecasters.} Within a fixed model, EPI rankings can still be reproducible; functional importance is therefore best treated as a property of the forecaster--task pair rather than a dataset-only graph.

\begin{figure}[!t]
\centering
\includegraphics[width=\linewidth]{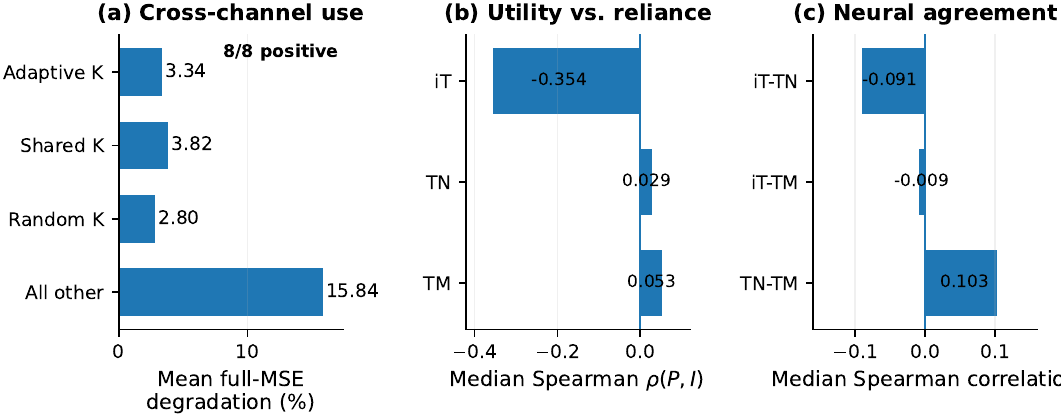}
\caption{\textbf{Functional source reliance.} (a) Grouped perturbations confirm substantial cross-channel use for iTransformer. (b) Controlled utility is weakly aligned with full-MSE reliance for iTransformer (iT), TimesNet (TN), and cross-channel TimeMixer (TM). (c) The three neural forecasters show little pairwise rank agreement under matched interventions.}
\label{fig:functional}
\end{figure}

\section{A Constructive Use of Cross-Channel Information}
\label{sec:constructive}
The preceding results identify a specific failure mode: externally defined channel structure can be informative under its defining mechanism yet become ineffective when imposed as the internal topology of a strong forecaster. We therefore test a deliberately weaker, model-compatible use of cross-channel information. A frozen PatchTST produces the channel-independent forecast $\widehat{\mathbf Y}^{\rm base}$ and patch latents $\mathbf z_{c,p}$. At each patch $p$, a lightweight support adapter applies channel self-attention, subtracts the original latent, and projects the difference into a bounded residual $\Delta_\theta(\mathbf X)$. The adapter output is limited to $0.5$ normalized units and modulated in 24-step forecast chunks by gates in $[0,1]$; the backbone remains frozen throughout. We then form
\begin{equation}
\widehat{\mathbf Y}^{\rm out}
=\widehat{\mathbf Y}^{\rm base}
+\alpha\Delta_\theta(\mathbf X),
\qquad 0\leq\alpha\leq1.
\label{eq:bounded-support}
\end{equation}
This is a residual adaptation of one frozen forecast, not averaging two independently trained forecasters; $\alpha=0$ exactly recovers the backbone and $\alpha=1$ applies the learned correction without post-hoc calibration. A global coefficient has a clipped closed-form least-squares solution on a temporally earlier calibration block. Channel-shrunk calibration estimates the same coefficient per channel and shrinks low-energy estimates toward the global value. Cross-fitted calibration estimates the global coefficient on each chronological half, retains it only when it improves the opposite half, and otherwise falls back to zero. No test label is used for training, calibration, or policy selection; Appendix~\ref{app:bounded-support} gives the exact estimators and locked protocol.

Table~\ref{tab:constructive} summarizes the locked evaluation over ETTh2, ETTm1, Exchange, and Weather, prediction lengths $H\in\{96,192,336,720\}$, and five seeds. Every support variant improves aggregate MSE over the frozen baseline. Cross-fitted calibration improves 12/16 dataset--horizon conditions, and none of the five-seed cell means regresses by more than $0.5\%$. Its overall gain is $0.388\%$, with a hierarchical bootstrap 95\% interval of $[0.155\%,0.661\%]$. The larger gain of uncalibrated support also shows that useful signal remains available, while the calibrated variants trade part of that gain for more conservative deployment. These results do not redefine $P$ or claim that the support path recovers neural reliance; instead, they show that cross-channel information can remain useful when introduced through an interaction compatible with the frozen forecaster.

\begin{table}[t]
\centering
\small
\setlength{\tabcolsep}{5pt}
\caption{\textbf{Locked post-hoc support evaluation.} Aggregate MSE gain is relative to the frozen channel-independent forecast; positive values are better. Confidence intervals use hierarchical bootstrap resampling over datasets, horizons, and seeds.}
\label{tab:constructive}
\begin{tabular}{lccc}
\toprule
Method & $\alpha$ policy & MSE gain & 95\% CI \\
\midrule
Uncalibrated support & fixed $1$ & $1.010\%$ & $[0.507,1.567]\%$ \\
Global calibration & global LS & $0.529\%$ & $[0.225,0.898]\%$ \\
Channel-shrunk calibration & channelwise shrinkage & $0.585\%$ & $[0.347,0.857]\%$ \\
Cross-fitted calibration & accepted two-fold LS & $0.388\%$ & $[0.155,0.661]\%$ \\
\bottomrule
\end{tabular}
\end{table}

\section{Discussion and Limitations}
\paragraph{The result is a separation, not a rejection of horizon adaptation.}
Related structure changes across future offsets, and Ridge-derived horizon-specific source sets can help both linear and nonlinear controlled predictors. The matched MLP result shows that the neural transfer gap is not explained simply by linear versus nonlinear prediction. Rather, $P$ is conditional on its source-selection protocol, and a neural forecaster can represent multivariate information through redundant, substitutable, or nonlinear combinations. The high-dimensional controls reinforce this interpretation by showing that aggregate Adaptive gains can vary with the evaluated target population and candidate-source pool.

\paragraph{Implication: adapt model-compatible interactions.}
Observed dependency $A$, controlled utility $P$, and functional reliance $I^{\mathcal F}$ answer different questions rather than estimating one hidden graph. Horizon-adaptive methods should therefore distinguish whether a relation changes, whether exploiting it improves prediction under a specified mechanism, and whether the trained forecaster actually relies on it. Section~\ref{sec:constructive} provides one concrete consequence: cross-channel information can be effective as a bounded, calibrated correction without being prescribed as internal topology. This does not make the three importance notions interchangeable; it instead avoids requiring such interchangeability.

\paragraph{Limitations and scope.}
The architecture-matched functional comparison covers three neural forecasters---iTransformer, TimesNet, and cross-channel TimeMixer---not the full space of multivariate models. EPI measures intervention sensitivity of a fixed trained model, not causal influence in the data-generating process. Individual-source analyses use selected targets and candidate pools in high dimensions, and the robustness controls show that controlled aggregate gains can remain sensitive to these choices. Accordingly, $P$ is an operational Ridge-based probe rather than a model-free population graph; the MLP experiment tests transfer of its selected sets rather than redefining $P$. Most forecasters are trained separately for each prediction length, while the fixed-$H=720$ analysis removes that checkpoint confound only for the complementary offset test. The bounded-support study establishes positive aggregate evidence rather than a formal non-degradation guarantee; its role here is to validate the design implication, not to introduce a new forecasting backbone.

\section{Conclusion}
We distinguish three notions of cross-channel importance: observed dependency, controlled predictive utility, and functional reliance. They need not coincide. Dependency changes across future offsets, horizon-specific source utility can improve controlled prediction, yet the resulting rankings do not reliably define a better neural topology or recover the sources used by trained forecasters. Moreover, matched interventions across three neural architectures reveal little agreement in functional source rankings. A locked post-hoc study nevertheless shows that external channel information can improve a frozen forecaster when introduced as bounded support rather than prescribed topology. Cross-channel structure should therefore be interpreted relative to the question and forecasting mechanism that define it, rather than as a single model-independent dataset graph: \textbf{related $\neq$ useful $\neq$ used}.

\section*{Reproducibility Statement}
The main text defines the three importance notions, primary interventions, and reported aggregate metrics. The appendix documents dataset protocols, complete controlled-selection results, the nonlinear MLP confirmation, high-dimensional target-subset/candidate-cap sensitivity, the 20-condition hard-mask study, alternative horizon-conditioning mechanisms, grouped and individual source interventions, same-checkpoint offset analysis, utility/candidate-pool robustness, EPI reliability, the TimesNet/TimeMixer confirmations, and the locked bounded-support protocol. The public reproducibility repository provides executable code and result summaries and is available at \url{https://github.com/dearyonghoon/rethinking-cross-channel}.

\clearpage
\appendix
% Consistent appendix table typography.
\setlength{\tabcolsep}{3.5pt}
\renewcommand{\arraystretch}{1.08}

\section{Experimental Setup and Reproducibility}
\label{app:protocol}
\label{app:details}

This section documents the configurations used throughout the paper. We distinguish three roles: (i) \emph{dependency-drift} experiments that measure how observed cross-channel neighborhoods vary across future offsets, (ii) \emph{controlled source-selection} experiments that estimate source utility with a transparent predictor, and (iii) \emph{functional-intervention} experiments that measure source reliance of fixed trained forecasters. Dependency estimation and source selection use training data only; held-out test outcomes are used only for final forecasting or post-training intervention evaluation.

\subsection{Datasets, splits, and sampling}

Controlled source-selection and forecasting experiments use history length $L=96$. PeMS03/04/07/08, Electricity, Weather, and Solar use chronological 70/10/20 train/validation/test splits for these forecasting experiments. ETTh1 uses 12/4/4 months at hourly resolution, and ETTm1 uses the same month counts at 15-minute resolution. The dependency-drift experiments use only a training region and are documented separately in Appendix~\ref{app:drift}; in particular, the PeMS drift study uses the first 60\% of each series, while the cross-domain drift study uses the standard training region. Each channel is standardized using its training-region mean and standard deviation, with a numerical floor of $10^{-6}$ on the standard deviation.

The controlled experiments use deterministic, evenly spaced origin subsampling when the number of valid origins exceeds the declared cap. High-dimensional datasets evaluate at most 32 evenly spaced target channels and at most 128 source candidates per target; the seven-channel ETT datasets use all channels.

\begin{table}[!ht]
\centering
\small
\begin{tabular}{lrrrrl}
\toprule
Dataset & Length & Channels & Split & $K_{\rm sel}$ & Prediction lengths \\
\midrule
PeMS03      & 26208 & 358 & 70/10/20 & 10 & 12, 24, 48 \\
PeMS04      & 16992 & 307 & 70/10/20 & 10 & 12, 24, 48 \\
PeMS07      & 28224 & 883 & 70/10/20 & 10 & 12, 24, 48 \\
PeMS08      & 17856 & 170 & 70/10/20 & 10 & 12, 24, 48 \\
Electricity & 26304 & 321 & 70/10/20 & 10 & 96, 192, 336, 720 \\
Weather     & 52696 & 21  & 70/10/20 & 10 & 96, 192, 336, 720 \\
Solar       & 52560 & 137 & 70/10/20 & 10 & 96, 192, 336, 720 \\
ETTh1       & 17420 & 7   & 12mo/4mo/4mo & 3 & 96, 192, 336, 720 \\
ETTm1       & 69680 & 7   & 12mo/4mo/4mo & 3 & 96, 192, 336, 720 \\
\bottomrule
\end{tabular}
\caption{Dataset geometry and final controlled-selection protocol. We use the dimension-aware rule $K_{\rm sel}(C)=\min(10,\lceil(C-1)/2\rceil)$, giving $K_{\rm sel}=3$ for ETTh1/ETTm1 and $10$ for the other datasets.}
\label{tab:app-datasets}
\end{table}

\begin{table}[!ht]
\centering
\small
\begin{tabular}{lrrr}
\toprule
Dataset & Evaluated targets & Train origins & Test origins \\
\midrule
PeMS03      & 32 & 5000 & 3000 \\
PeMS04      & 32 & 5000 & 3000 \\
PeMS07      & 32 & 5000 & 3000 \\
PeMS08      & 32 & 5000 & 3000 \\
Electricity & 32 & 5000 & 3000 \\
Weather     & 21 & 5000 & 3000 \\
Solar       & 32 & 5000 & 3000 \\
ETTh1       & 7  & 5000 & 2161 \\
ETTm1       & 7  & 5000 & 3000 \\
\bottomrule
\end{tabular}
\caption{Effective sampling sizes in the final controlled source-selection experiment, taken from the executed rolling-stability notebook after valid-origin construction and deterministic subsampling.}
\label{tab:app-sampling}
\end{table}

For high-dimensional datasets, the candidate pool is horizon-invariant and capped at 128 sources using training-only absolute current-value correlation. The dimension-aware source budget above is used instead of a fixed $K=10$ so that ETTh1 and ETTm1 remain genuinely sparse.

\subsection{Controlled predictive-utility protocol}

The controlled predictor summarizes each target or source history by four features: the last value, 3-step mean, 12-step mean, and 12-step change, and predicts the target endpoint at $t+H$. Within the official training region, the first 70\% of sampled origins fit the target-only model and source corrections; the remaining 30\% score source utility. The target-only and source-correction Ridge coefficients are $10^{-3}$ and $10^{-2}$, respectively. Candidate-source features are linearly residualized with respect to the target-history features before fitting the source correction, so the measured source utility is incremental to information already represented by the target history.

All channels are evaluated when $C\leq32$; otherwise 32 approximately evenly spaced target channels are used. The train/test origin caps are 5,000/3,000. All controlled experiments use seed 2026.

\subsection{Phase-1 neural backbones}

The external backbone benchmark uses seed 2026, history length 96, label length 48, prediction lengths 96/192/336/720, at most 30 epochs, and early-stopping patience 6 based on validation MSE. Models use Adam and gradient-norm clipping at 1.0. PatchTST keeps its base learning rate for two epochs and then applies a 0.9 multiplicative decay per epoch; the other baselines halve the learning rate every five epochs.

\begin{table}[!ht]
\centering
\small
\setlength{\tabcolsep}{3pt}
\begin{tabular}{lrrrp{0.40\textwidth}}
\toprule
Model & Parameters & Base LR & Base batch & Configuration \\
\midrule
DLinear & 18,624 & $5{\times}10^{-3}$ & 32 &
individual=False, moving avg.=25 \\
PatchTST & 547,296 & $10^{-4}$ & 32 &
$e=3$, heads=16, $d=128$, ff=256, patch=16, stride=8, RevIN \\
TimesNet & 2,363,783 & $10^{-4}$ & 32 &
$e=2$, $d=32$, ff=64, heads=8, top-$k=5$, kernels=6, factor=3 \\
iTransformer & 4,833,888 & $10^{-4}$ & 32 &
$e=3$, heads=8, $d=512$, ff=512, factor=3 \\
\bottomrule
\end{tabular}
\caption{Phase-1 forecasting backbones. Electricity uses batch size at most 16 for all models, with iTransformer further reduced to 8 for $H\geq336$; Solar uses batch size at most 16 for all models. PatchTST uses dropout/fc-dropout 0.2/0.2, head dropout 0, padding-to-end, and RevIN with affine disabled. TimesNet and iTransformer use dropout 0.1.}
\label{tab:app-model-config}
\end{table}

\subsection{Functional-intervention and reliability protocol}

The final reliability analysis uses three preselected conditions: Electricity $H=192$, Solar $H=720$, and ETTh1 $H=720$. Candidate pools are inherited from the preceding forecaster-specific robustness experiment. For the high-dimensional datasets, the base pool combines the top 8 training-only current-correlation sources with up to 12 consensus predictive-teacher sources aggregated across the four standard prediction lengths, after which 10 deterministic random sources not already in the pool are added. ETTh1 uses all six non-target channels.

\begin{table}[!ht]
\centering
\small
\begin{tabular}{lrrrr}
\toprule
Condition & Targets & Unique sources & Target-source pairs & Test windows \\
\midrule
Electricity, $H=192$ & 6 & 132 & 173 & 512 \\
Solar, $H=720$       & 6 & 94  & 178 & 512 \\
ETTh1, $H=720$       & 7 & 7   & 42  & 512 \\
\midrule
Total                 & 19 & -- & 393 & -- \\
\bottomrule
\end{tabular}
\caption{Final EPI evaluation size. ``Unique sources'' counts the union of source indices appearing across selected targets; target-source pairs exclude self-pairs.}
\label{tab:app-epi-size}
\end{table}

Each condition uses 512 test windows and 16 common donor mappings for every source. Donors are nonzero cyclic permutations sampled without replacement from shifts whose circular displacement is at least one eighth of the evaluation set, using the deterministic seed defined in the final EPI notebook. The same donor mappings are applied to every source within a condition, preserving each donor channel's complete temporal history while breaking alignment with the current target context.

Reliability is measured with 50 donor split-half repetitions (8 donors versus 8 donors) and 50 random-window split-half repetitions. Temporal stability uses four contiguous blocks of the 512 evaluation windows. Full-horizon MSE is the primary functional-reliance metric, full-horizon MAE is the robustness metric, and endpoint MSE is secondary.

The TimesNet confirmation reuses exactly the iTransformer target--source pairs, test origins, and 16 donor mappings. It uses the Phase-1 TimesNet configuration in Table~\ref{tab:app-model-config}; inference batch sizes are 8, 8, and 32 for Electricity, Solar, and ETTh1, respectively. No TimesNet hyperparameter, candidate pool, donor mapping, or reliability threshold is tuned after observing the confirmatory results.

The third-backbone confirmation uses TimeMixer with explicit cross-channel mixing (\texttt{channel\_independence=0}), two mixing layers, $d_{\rm model}=32$, $d_{\rm ff}=32$, dropout 0.1, moving-average window 25, and three average-pooling down-sampling layers with window 2. It is trained with Adam at learning rate $10^{-3}$, halved every five epochs, for at most 20 epochs with patience 5. TimeMixer reuses the same three conditions, target--source pairs, 512 test origins, and 16 donor mappings as iTransformer and TimesNet.

\section{Full Cross-Offset Dependency-Drift Results}
\label{app:drift}

This section expands the dependency-drift analysis from the main paper. We keep the absolute training-only cross-lag correlation $A_{ij}(h)$ defined in the main text and focus here on the stability control and the complete empirical results. For a target-specific Top-$K$ neighborhood $\mathcal N_{i,K}(h)$, the primary statistic subtracts same-offset split instability from the observed cross-offset neighborhood change:
\[
D^{\rm adj}_{i,K}(h_a,h_b)
=
\bigl[1-J(\mathcal N_{i,K}(h_a),\mathcal N_{i,K}(h_b))\bigr]
-\frac{1}{2}\sum_{h\in\{h_a,h_b\}}
\bigl[1-J(\mathcal N^{A}_{i,K}(h),\mathcal N^{B}_{i,K}(h))\bigr].
\]
A positive value therefore means that the neighborhood changes across future offsets more than would be expected from same-offset estimation instability alone. The sparse-neighborhood statistic is primary; full-ranking Spearman drift is reported as a secondary diagnostic because the full ranking contains many weak source channels.

\subsection{Traffic datasets}

The PeMS drift study uses only the first 60\% of each series. Split-half stability is estimated by assigning alternating one-day blocks (288 five-minute samples) to split A or B. We use fixed $K\in\{10,20\}$. For prediction lengths $H\in\{12,24,48\}$, the relative future anchors are $[1,3,6,12]$, $[1,6,12,24]$, and $[1,12,24,48]$, respectively.

Table~\ref{tab:app-pems-drift} reports both endpoint drift, comparing $h=1$ with $h=H$, and the average over all six anchor pairs within each prediction length. Under the primary $K=20$ setting, all four datasets exhibit larger adjusted drift as the prediction length increases. At $H=48$, endpoint adjusted drift ranges from 0.622 on PeMS08 to 0.726 on PeMS03.

\begin{table}[!ht]
\centering
\small
\begin{minipage}[t]{0.47\linewidth}
\centering
\textbf{Endpoint adjusted drift, $h=1$ vs. $h=H$}\\[2pt]
\begin{tabular}{lrrr}
\toprule
Dataset & $H=12$ & $H=24$ & $H=48$ \\
\midrule
PeMS03 & 0.4716 & 0.6359 & 0.7257 \\
PeMS04 & 0.3171 & 0.5522 & 0.6408 \\
PeMS07 & 0.5180 & 0.6608 & 0.6801 \\
PeMS08 & 0.2594 & 0.5109 & 0.6217 \\
\bottomrule
\end{tabular}
\end{minipage}\hfill
\begin{minipage}[t]{0.47\linewidth}
\centering
\textbf{Mean adjusted drift over all anchor pairs}\\[2pt]
\begin{tabular}{lrrr}
\toprule
Dataset & $H=12$ & $H=24$ & $H=48$ \\
\midrule
PeMS03 & 0.2167 & 0.4161 & 0.5409 \\
PeMS04 & 0.0820 & 0.3049 & 0.4632 \\
PeMS07 & 0.2455 & 0.4464 & 0.5765 \\
PeMS08 & 0.0656 & 0.2528 & 0.4077 \\
\bottomrule
\end{tabular}
\end{minipage}
\caption{PeMS noise-adjusted Top-20 dependency drift. The left panel reports endpoint drift by comparing the first and last future offsets for each prediction length; the right panel averages adjusted drift over all six pairs among the four relative anchors.}
\label{tab:app-pems-drift}
\end{table}

\begin{figure}[!t]
\centering
\includegraphics[width=0.60\textwidth]{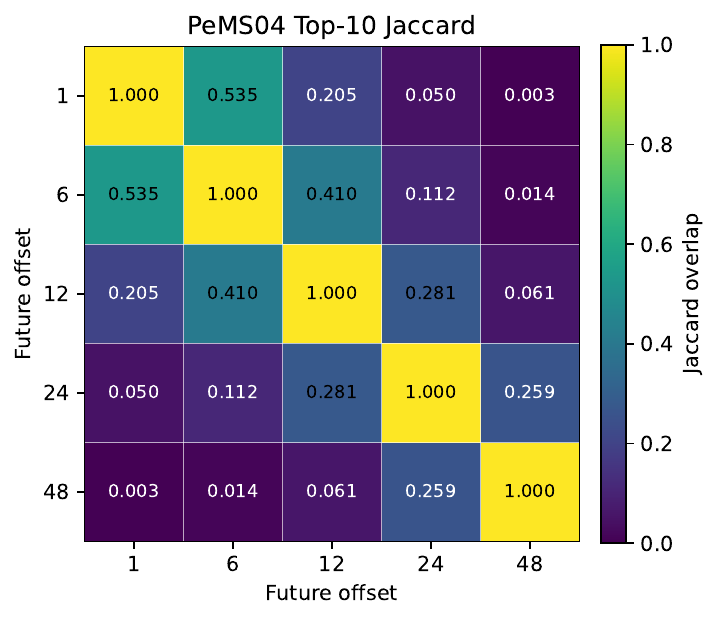}
\caption{\textbf{PeMS04 Top-10 neighborhood overlap.} Mean target-wise Jaccard similarity decreases sharply as the future offsets separate. In particular, $J(1,48)=0.0031$ and $J(12,48)=0.0612$. This figure shows raw overlap for interpretability; the primary statistical analysis uses the noise-adjusted drift above.}
\label{fig:app-pems04}
\end{figure}

PeMS04 provides a particularly transparent raw-overlap example. Its Top-10 neighborhood at $h=1$ is almost completely replaced by $h=48$: the mean Jaccard overlap is 0.0031 and the median is zero. The same pattern appears progressively as the offset separation grows (Fig.~\ref{fig:app-pems04} and Table~\ref{tab:app-pems04-pairs}).

\begin{table}[!ht]
\centering
\small
\begin{tabular}{rrrrrr}
\toprule
$h_a$ & $h_b$ & Mean J & Median J & Q25 & Q75 \\
\midrule
1 & 6  & 0.5351 & 0.5385 & 0.3333 & 0.6667 \\
1 & 12 & 0.2054 & 0.1765 & 0.0526 & 0.3333 \\
1 & 24 & 0.0504 & 0.0000 & 0.0000 & 0.0526 \\
1 & 48 & 0.0031 & 0.0000 & 0.0000 & 0.0000 \\
6 & 12 & 0.4103 & 0.3333 & 0.2500 & 0.5385 \\
6 & 24 & 0.1120 & 0.0526 & 0.0000 & 0.1111 \\
6 & 48 & 0.0144 & 0.0000 & 0.0000 & 0.0000 \\
12 & 24 & 0.2811 & 0.2500 & 0.0526 & 0.4286 \\
12 & 48 & 0.0612 & 0.0000 & 0.0000 & 0.0526 \\
24 & 48 & 0.2587 & 0.1765 & 0.0526 & 0.4286 \\
\bottomrule
\end{tabular}
\caption{PeMS04 target-wise Top-10 Jaccard summaries across future offsets.}
\label{tab:app-pems04-pairs}
\end{table}

\subsection{Cross-domain results}

The cross-domain study uses the standard training region: the first 70\% for Electricity, Weather, and Solar, and the first 12 months for ETTh1/ETTm1. Same-offset stability is estimated from alternating one-week blocks. For each prediction length $H\in\{96,192,336,720\}$, we evaluate four anchors: $h=1$, approximately $H/4$, approximately $H/2$, and $H$. The primary sparse budget uses 10\% of the available non-target channels, floored at 2 and capped at 20; a 5\% budget provides a robustness check. Confidence intervals are obtained by 2,000 bootstrap resamples of target channels with seed 2027.

Table~\ref{tab:app-drift-full} gives the complete primary-budget endpoint results. The 95\% confidence interval is strictly above zero in 18 of 20 dataset--prediction-length conditions. The two exceptions are Weather at $H=720$ and ETTm1 at $H=96$. Electricity and Solar show especially high positive-target fractions, while the seven-channel ETT datasets exhibit coarser target-level fractions because only seven targets and a two-source neighborhood are available.

\begin{figure}[!t]
\centering
\includegraphics[width=0.82\textwidth]{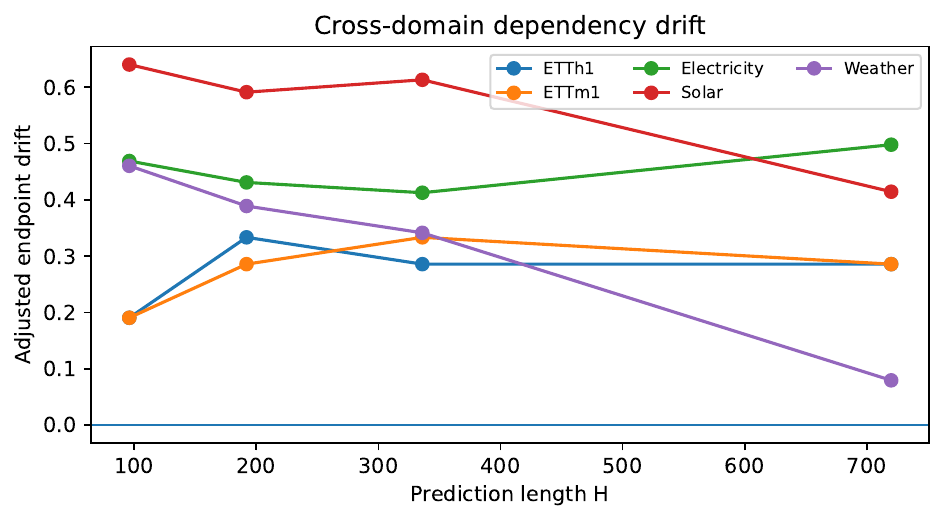}
\caption{\textbf{Cross-domain adjusted endpoint drift.} Horizon-dependent cross-channel drift is broad but not universally monotonic in prediction length. No evaluated non-traffic dataset exhibits a monotonic non-decreasing trajectory over all four prediction lengths under the primary sparse metric.}
\label{fig:app-drift-general}
\end{figure}

\begin{table}[!ht]
\centering
\small
\begin{tabular}{lrrlrr}
\toprule
Dataset & $H$ & $K$ & Adjusted drift [95\% CI] & Positive targets & Adj. rank drift \\
\midrule
Electricity & 96  & 20 & 0.4688 [0.4460, 0.4924] & 0.9657 & 0.1119 \\
Electricity & 192 & 20 & 0.4308 [0.4080, 0.4506] & 0.9657 & 0.0820 \\
Electricity & 336 & 20 & 0.4125 [0.3892, 0.4367] & 0.9564 & 0.0701 \\
Electricity & 720 & 20 & 0.4978 [0.4771, 0.5172] & 0.9938 & 0.1251 \\
Weather & 96  & 2 & 0.4603 [0.2619, 0.6429] & 0.7619 & 0.1159 \\
Weather & 192 & 2 & 0.3889 [0.1984, 0.5635] & 0.7619 & 0.1140 \\
Weather & 336 & 2 & 0.3413 [0.1270, 0.5635] & 0.7143 & 0.1046 \\
Weather & 720 & 2 & 0.0794 [-0.0238, 0.1905] & 0.3810 & -0.0453 \\
Solar & 96  & 14 & 0.6401 [0.6052, 0.6701] & 0.9708 & 0.6496 \\
Solar & 192 & 14 & 0.5910 [0.5589, 0.6224] & 0.9854 & 0.6800 \\
Solar & 336 & 14 & 0.6130 [0.5799, 0.6421] & 0.9927 & 0.6884 \\
Solar & 720 & 14 & 0.4143 [0.3816, 0.4452] & 0.9635 & 0.4518 \\
ETTh1 & 96  & 2 & 0.1905 [0.0476, 0.3810] & 0.4286 & 0.0571 \\
ETTh1 & 192 & 2 & 0.3333 [0.0952, 0.5714] & 0.5714 & 0.1959 \\
ETTh1 & 336 & 2 & 0.2857 [0.0952, 0.5714] & 0.4286 & 0.2408 \\
ETTh1 & 720 & 2 & 0.2857 [0.0952, 0.4762] & 0.5714 & 0.4041 \\
ETTm1 & 96  & 2 & 0.1905 [0.0000, 0.3810] & 0.2857 & 0.0857 \\
ETTm1 & 192 & 2 & 0.2857 [0.0952, 0.5714] & 0.4286 & 0.0041 \\
ETTm1 & 336 & 2 & 0.3333 [0.0952, 0.5714] & 0.5714 & 0.3306 \\
ETTm1 & 720 & 2 & 0.2857 [0.0952, 0.4762] & 0.5714 & 0.4367 \\
\bottomrule
\end{tabular}
\caption{Cross-domain adjusted endpoint dependency drift under the primary 10\% sparse-budget policy. ``Positive targets'' is the fraction of target channels with positive adjusted drift. Adjusted rank drift is reported as a secondary full-ranking diagnostic.}
\label{tab:app-drift-full}
\end{table}

The secondary rank-based diagnostic reinforces that sparse-neighborhood drift and full-ranking drift need not behave identically. Electricity, for example, has substantial sparse endpoint drift (mean 0.453 across prediction lengths) while its mean adjusted full-ranking drift is only 0.097. This supports treating the sparse neighborhood as a distinct structural object rather than interpreting every ranking change as equivalent.

\subsection{Robustness to the sparse budget}

The 5\% and 10\% policies differ only for Electricity and Solar; Weather and both ETT datasets are floored at $K=2$ under both policies. Table~\ref{tab:app-drift-budget} therefore reports the nontrivial cases. All eight Electricity/Solar conditions retain positive adjusted drift with confidence intervals above zero under both budgets. The numerical magnitude changes, particularly for Solar at $H=96$, but the qualitative conclusion is unchanged.

\begin{table}[!ht]
\centering
\small
\begin{tabular}{lrrrl}
\toprule
Dataset & $H$ & $K_{5\%}$ / drift & $K_{10\%}$ / drift & 95\% CI $>0$ under both? \\
\midrule
Electricity & 96  & 16 / 0.4804 & 20 / 0.4688 & yes \\
Electricity & 192 & 16 / 0.4422 & 20 / 0.4308 & yes \\
Electricity & 336 & 16 / 0.4223 & 20 / 0.4125 & yes \\
Electricity & 720 & 16 / 0.5193 & 20 / 0.4978 & yes \\
Solar & 96  & 7 / 0.8023 & 14 / 0.6401 & yes \\
Solar & 192 & 7 / 0.5606 & 14 / 0.5910 & yes \\
Solar & 336 & 7 / 0.5724 & 14 / 0.6130 & yes \\
Solar & 720 & 7 / 0.3778 & 14 / 0.4143 & yes \\
\bottomrule
\end{tabular}
\caption{Sparse-budget robustness where the 5\% and 10\% policies produce different neighborhood sizes. The table reports endpoint adjusted drift; both policies have strictly positive 95\% bootstrap lower bounds in all listed conditions.}
\label{tab:app-drift-budget}
\end{table}

\subsection{Summary}

The traffic and cross-domain studies support two deliberately separate conclusions. First, cross-offset neighborhood changes are not explained solely by same-offset estimation instability: all PeMS conditions exhibit substantial positive adjusted endpoint drift, and 18 of 20 non-traffic conditions have a strictly positive bootstrap lower bound. Second, the amount of drift is domain dependent. PeMS shows a clear increase with prediction length, whereas the general benchmarks do not exhibit a universal monotonic trend. The main paper therefore treats \emph{the existence of horizon-dependent cross-channel structure} as the robust empirical finding, while avoiding the stronger claim that dependency drift must increase monotonically with prediction length.

\section{Complete Controlled Source-Selection Results}
\label{app:controlled}

This section reports the full controlled comparison between a \emph{shared} source set and a \emph{horizon-adaptive} source set. Both policies use exactly the same target-history features, source features, residualization procedure, Ridge predictor, train/test origins, and source budget described in Appendix~\ref{app:protocol}. They differ only in source selection: Shared ranks sources by predictive utility averaged across the prediction lengths of a dataset, whereas Adaptive ranks sources separately for each prediction length. Source utilities are estimated using training data only; the final comparison is evaluated on held-out test origins.

For each dataset--prediction-length condition, we report the pooled endpoint-MSE gain of Adaptive over Shared,
\[
G_{\rm adapt}(H)
=
100\,\frac{\sum_i \mathrm{MSE}^{\rm shared}_i(H)-\sum_i \mathrm{MSE}^{\rm adapt}_i(H)}
{\sum_i \mathrm{MSE}^{\rm shared}_i(H)}.
\]
Positive values therefore mean that changing the selected source channels with the prediction length improves the controlled predictor. Pooling is performed over the evaluated target channels before taking the percentage, rather than averaging target-wise percentage gains.

\subsection{General forecasting datasets}

\begin{figure}[!t]
\centering
\includegraphics[width=0.72\textwidth]{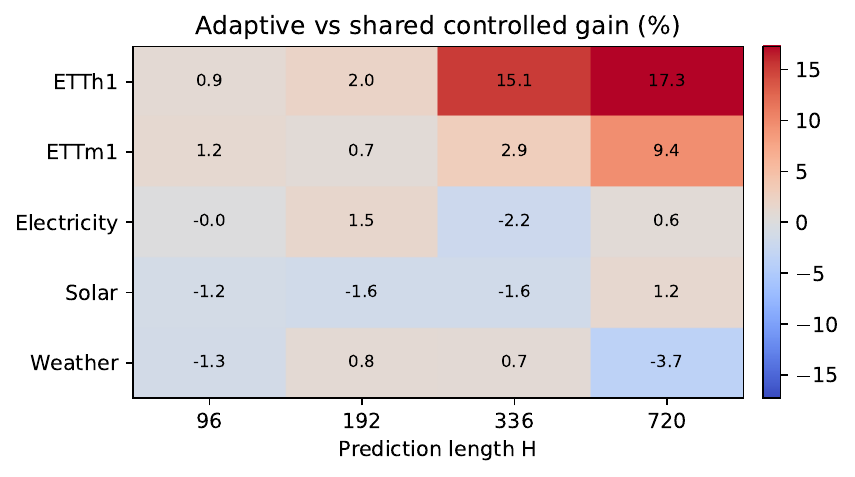}
\caption{\textbf{Controlled Adaptive-versus-Shared gain on the five general forecasting datasets.} Positive cells favor horizon-specific source selection. The ETT datasets are positive at all four prediction lengths, whereas Electricity, Solar, and Weather remain mixed.}
\label{fig:app-controlled-general}
\end{figure}

Figure~\ref{fig:app-controlled-general} and Table~\ref{tab:app-controlled-general} give all 20 conditions from Electricity, Weather, Solar, ETTh1, and ETTm1. Adaptive selection improves 13 of the 20 conditions. The mean gain across these conditions is $+2.137\%$ and the median is $+0.770\%$.

\begin{table}[!ht]
\centering
\small
\begin{tabular}{lrrrr}
\toprule
Dataset & $H=96$ & $H=192$ & $H=336$ & $H=720$ \\
\midrule
ETTh1       & 0.8783  & 1.9576  & 15.1233 & 17.2641 \\
ETTm1       & 1.1934  & 0.6612  & 2.9182  & 9.4491 \\
Electricity & -0.0445 & 1.4992  & -2.1514 & 0.6137 \\
Solar       & -1.1805 & -1.5726 & -1.5895 & 1.2143 \\
Weather     & -1.3256 & 0.8131  & 0.7264  & -3.7116 \\
\midrule
Mean        & -0.0958 & 0.6717  & 3.0054  & 4.9659 \\
\bottomrule
\end{tabular}
\caption{Controlled Adaptive-versus-Shared endpoint-MSE gain (\%) on the five general forecasting datasets. The bottom row averages the five datasets at each prediction length.}
\label{tab:app-controlled-general}
\end{table}

The aggregate tendency becomes more favorable at longer prediction lengths, with mean gains of $-0.096$, $+0.672$, $+3.005$, and $+4.966\%$ at $H=96,192,336,720$, respectively. We do not interpret this as a universal monotonic law: individual datasets are clearly non-monotonic. Instead, the result shows that horizon-specific source utility can be materially useful in some regimes, especially the longer-horizon ETT conditions, while being neutral or harmful in others.

The corrected low-dimensional setting is important here. Under the final dimension-aware source budget, ETTh1 and ETTm1 use $K_{\rm sel}=3$ rather than the earlier fixed $K=10$ screen, which would have selected all six non-target channels. The final experiment therefore tests genuine sparse source selection. Across the ETT target--horizon evaluations, the mean Shared--Adaptive source-set Jaccard is 0.6964, confirming that the two policies are not identical. At the same time, only 28.57\% of individual ETT target--horizon cases have lower Adaptive MSE than Shared, despite positive pooled gains in all eight dataset-level conditions. This heterogeneity is one reason we avoid interpreting a positive pooled gain as evidence that every target benefits from adaptation.

\subsection{Traffic datasets}

The PeMS results are more strongly horizon dependent. Adaptive selection is positive in 8 of 12 conditions overall, with a mean gain of $+10.191\%$ and a median gain of $+6.559\%$. At the longest prediction length, $H=48$, all four datasets favor Adaptive, with a mean gain of $+26.474\%$.

\begin{figure}[!t]
\centering
\includegraphics[width=0.70\textwidth]{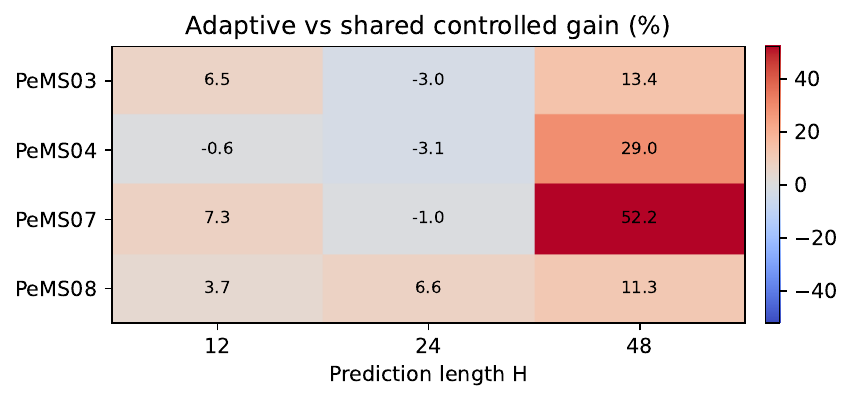}
\caption{\textbf{Controlled Adaptive-versus-Shared gain on PeMS.} The longest prediction length is positive on all four datasets, with particularly large gains on PeMS04 and PeMS07.}
\label{fig:app-controlled-pems}
\end{figure}

\begin{table}[!ht]
\centering
\small
\begin{tabular}{lrrr}
\toprule
Dataset & $H=12$ & $H=24$ & $H=48$ \\
\midrule
PeMS03 & 6.4896  & -2.9622 & 13.4069 \\
PeMS04 & -0.5556 & -3.0974 & 28.9567 \\
PeMS07 & 7.2736  & -1.0412 & 52.2475 \\
PeMS08 & 3.6561  & 6.6286  & 11.2838 \\
\midrule
Mean   & 4.2159  & -0.1181 & 26.4737 \\
\bottomrule
\end{tabular}
\caption{Controlled Adaptive-versus-Shared endpoint-MSE gain (\%) on PeMS.}
\label{tab:app-controlled-pems}
\end{table}

The PeMS pattern is also non-monotonic at intermediate prediction lengths: the mean gain is $+4.216\%$ at $H=12$, $-0.118\%$ at $H=24$, and $+26.474\%$ at $H=48$. Thus, even in the family where the longest-horizon effect is strongest, simply increasing the prediction length does not guarantee that horizon-specific source selection will help.

\subsection{Aggregate result and interpretation}

Across all 32 dataset--prediction-length conditions, Adaptive is better than Shared in 21 conditions. The mean pooled gain is $+5.157\%$ and the median is $+1.036\%$. These results establish the specific claim needed by the main paper: \emph{horizon-specific source utility can be predictively actionable under a controlled predictor}. They do not establish that Adaptive selection is universally better, nor that the same source topology should be imposed on a separately trained neural forecaster. Appendix~\ref{app:gating} next examines whether the profitable controlled cases can be identified reliably from historical temporal stability alone.

\subsection{Nonlinear controlled-predictor confirmation}
\label{app:mlp-controlled}
To test whether the controlled result is specific to a linear final predictor, we keep the Ridge-derived Shared and Adaptive source sets fixed and replace only the final forecaster with a small MLP. The MLP uses two 128-unit hidden layers with GELU activations, matched initialization and mini-batch order for Shared and Adaptive, a chronological 80/20 inner validation split, AdamW, and early stopping. We repeat the full experiment with seeds 2026, 2027, and 2028.

\begin{table}[!ht]
\centering
\small
\begin{tabular}{lrrrrr}
\toprule
Scope & Wins & Mean gain & Median gain & $\rho$(Ridge, MLP) & Sign agree\\
\midrule
General 20 & 13/20 & +2.137 & +1.852 & 0.713 & 16/20\\
All 32 & 19/32 & +1.667 & +1.154 & 0.472 & 22/32\\
\bottomrule
\end{tabular}
\caption{Matched nonlinear controlled-predictor confirmation. Gains are Adaptive-versus-Shared pooled endpoint-MSE reductions (\%). The General-20 row is the condition set shared with the iTransformer hard-mask experiment.}
\label{tab:app-mlp-confirm}
\end{table}

The result is stable across MLP seeds: the 32-condition win counts are 21, 18, and 19 for seeds 2026, 2027, and 2028, with mean gains $+1.354\%$, $+1.896\%$, and $+1.729\%$, respectively. The main paper focuses on the General-20 subset because it enables a matched comparison with iTransformer. There, Ridge and MLP gains are strongly rank-aligned ($\rho=0.713$), which argues against interpreting the utility--topology mismatch as a trivial consequence of comparing a linear controlled predictor with a nonlinear neural forecaster. The weaker all-32 alignment reflects more heterogeneous PeMS behavior and is reported to keep the scope explicit.

\subsection{High-dimensional target-subset and candidate-cap sensitivity}
\label{app:highdim-robustness}
The primary controlled study uses a deterministic 32-target subset when $C>32$ and a training-only correlation screen capped at 128 candidate sources. To test whether those computational choices define the conclusion, we repeat the complete pooled endpoint-MSE comparison on Electricity and PeMS04 with five predeclared random 32-target subsets (seeds 2026--2030), and separately vary the candidate cap over 64, 128, and 192 while holding the original target subset fixed. The original cap-128 results are reproduced to within $0.004$ percentage points.

\begin{table}[!ht]
\centering
\small
\setlength{\tabcolsep}{3.6pt}
\begin{tabular}{lrrrrr}
\toprule
Dataset & $H$ & Mean $\pm$ std & Min & Max & Positive seeds \\
\midrule
Electricity & 96  & $0.206\pm4.889$ & -5.061 & 5.520 & 3/5 \\
Electricity & 192 & $2.006\pm1.216$ & 0.495 & 3.811 & 5/5 \\
Electricity & 336 & $-1.405\pm3.475$ & -5.024 & 3.778 & 2/5 \\
Electricity & 720 & $-1.711\pm2.646$ & -4.033 & 2.782 & 1/5 \\
PeMS04 & 12 & $2.169\pm2.882$ & -0.193 & 5.322 & 3/5 \\
PeMS04 & 24 & $-0.034\pm1.730$ & -2.641 & 1.724 & 3/5 \\
PeMS04 & 48 & $21.697\pm16.097$ & -2.663 & 36.681 & 4/5 \\
\bottomrule
\end{tabular}
\caption{Random-target robustness for the controlled Adaptive-versus-Shared experiment. Entries are pooled endpoint-MSE gains (\%) across five independently sampled 32-target subsets at the original 128-source candidate cap.}
\label{tab:app-targetsubset}
\end{table}

\begin{table}[!ht]
\centering
\small
\setlength{\tabcolsep}{5pt}
\begin{tabular}{lrrrr}
\toprule
Dataset & $H$ & Cap 64 & Cap 128 & Cap 192 \\
\midrule
Electricity & 96  & 0.391 & -0.045 & 0.557 \\
Electricity & 192 & 2.112 & 1.499 & 1.055 \\
Electricity & 336 & 1.343 & -2.151 & -0.550 \\
Electricity & 720 & 1.439 & 0.614 & -0.315 \\
PeMS04 & 12 & -0.257 & -0.552 & -0.484 \\
PeMS04 & 24 & -1.131 & -3.098 & -5.987 \\
PeMS04 & 48 & -8.839 & 28.958 & 13.374 \\
\bottomrule
\end{tabular}
\caption{Candidate-cap sensitivity on the original deterministic target subset. Entries are pooled Adaptive-versus-Shared endpoint-MSE gains (\%). Several conditions change sign as the admissible candidate pool expands.}
\label{tab:app-candidatecap}
\end{table}

The robustness result is deliberately interpreted as a scope qualifier rather than a new performance claim. Electricity $H=192$ remains positive for all five target subsets and all three candidate caps, while PeMS04 $H=48$ is positive for four of five target subsets but changes sign across candidate caps. Other conditions are mixed. Thus the predeclared cap-128 experiment remains a valid controlled comparison, but its aggregate gain should be understood as conditional on the evaluated target population and admissible source pool rather than as an intrinsic dataset-level quantity.

\section{Selective Routing and Temporal-Stability Diagnostics}
\label{app:gating}

The controlled results show that horizon-adaptive source selection can improve forecasting, but the benefit is not universal across targets and prediction lengths. We therefore ask whether training-only temporal stability can identify when Adaptive selection should be used. The official training region is divided into five chronological blocks, which induce four expanding-window historical evaluations of the Adaptive-versus-Shared gain. The rolling gate selects Adaptive only when the median historical gain is positive and at least three of the four historical gains are positive; otherwise, it selects Shared. As a simpler diagnostic, the Last-window gate selects Adaptive only when the most recent historical gain is positive. Oracle selects between Adaptive and Shared using test outcomes and is included only as an unattainable upper bound.

\begin{table}[!ht]
\centering
\small
\begin{tabular}{lrrrrrr}
\toprule
Family & Cond. & Mean roll. & Median roll. & Mean last & Mean adapt. & Mean oracle \\
\midrule
General & 20 & -0.1493 & 0.0000 & 0.3608 & 2.1368 & 3.9203 \\
PeMS & 12 & 5.7509 & 0.2146 & 8.5692 & 10.1905 & 18.7229 \\
\bottomrule
\end{tabular}
\vspace{2mm}

\begin{tabular}{lrrr}
\toprule
Family & Roll. positive & Roll. $>$ adapt. & Roll. $>$ last \\
\midrule
General & 0.3000 & 0.3000 & 0.3500 \\
PeMS & 0.5833 & 0.4167 & 0.3333 \\
\bottomrule
\end{tabular}
\caption{Summary of the temporal-stability routing diagnostics. Top: mean and median test gains (\%) for the rolling gate, together with the mean gains of the Last-window gate, always-Adaptive selection, and Oracle. Bottom: fraction of conditions in which the rolling gate has positive test gain, outperforms always-Adaptive selection, or outperforms the Last-window gate. Oracle uses test outcomes only and is not a deployable policy.}
\label{tab:app-gate-summary}
\end{table}

The rolling gate is not a reliable improvement over always-Adaptive selection. Across the 20 general conditions, its mean gain is $-0.1493\%$, whereas always-Adaptive selection averages $+2.1368\%$ and Oracle averages $+3.9203\%$. On PeMS, the rolling gate averages $+5.7509\%$, but always-Adaptive selection is stronger on average ($+10.1905\%$), with substantial remaining Oracle headroom ($+18.7229\%$).

Historical temporal stability is also only weakly associated with the realized benefit of Adaptive selection. Across the 780 evaluated dataset--target--horizon instances, the Spearman correlations between the final Adaptive-versus-Shared test gain and the historical rolling positive rate, rolling median gain, and most recent historical gain are approximately 0.018, 0.067, and 0.166, respectively. We therefore do not adopt the temporal-stability gate in the main analysis. This negative result distinguishes two questions that need not have the same answer: horizon-specific source selection can be predictively useful under the controlled predictor, yet simple historical persistence may still be insufficient to determine reliably when that adaptation should be activated.

\section{Phase-1 Forecasting Benchmark and Backbone Selection}
\label{app:phase1}

Before interpreting a failure to transfer controlled source structure, we verified that the analysis was not tied to a weak custom backbone. DLinear, PatchTST, TimesNet, and iTransformer are trained in a common dependency-free data pipeline, and Hybrid-0 is retained only as a record of the earlier prototype stage.

\begin{table}[!ht]
\centering
\small
\begin{tabular}{lrrrrrr}
\toprule
Dataset & H & DLinear & Hybrid-0 & PatchTST & TimesNet & iTransformer \\
\midrule
ETTh1 & 96 & 0.3887 & 0.4116 & 0.3824 & 0.3965 & 0.3944 \\
ETTh1 & 192 & 0.4516 & 0.4895 & 0.4295 & 0.4509 & 0.4339 \\
ETTh1 & 336 & 0.5029 & 0.5995 & 0.4706 & 0.5500 & 0.4769 \\
ETTh1 & 720 & 0.5412 & 0.7084 & 0.4860 & 0.5427 & 0.4859 \\
ETTm1 & 96 & 0.3428 & 0.3809 & 0.3331 & 0.3284 & 0.3442 \\
ETTm1 & 192 & 0.3868 & 0.4278 & 0.3716 & 0.3966 & 0.3775 \\
ETTm1 & 336 & 0.4163 & 0.4889 & 0.4041 & 0.4260 & 0.4279 \\
ETTm1 & 720 & 0.4926 & 0.5464 & 0.4618 & 0.5324 & 0.5039 \\
Electricity & 96 & 0.1948 & 0.1907 & 0.1968 & 0.1800 & 0.1513 \\
Electricity & 192 & 0.1941 & 0.2004 & 0.2058 & 0.1954 & 0.1643 \\
Electricity & 336 & 0.2070 & 0.2164 & 0.2241 & 0.2129 & 0.1782 \\
Electricity & 720 & 0.2475 & 0.2535 & 0.2641 & 0.2459 & 0.2114 \\
Solar & 96 & 0.2853 & 0.1984 & 0.2414 & 0.2235 & 0.2034 \\
Solar & 192 & 0.3180 & 0.2186 & 0.2813 & 0.2721 & 0.2410 \\
Solar & 336 & 0.3493 & 0.2442 & 0.2902 & 0.2947 & 0.2543 \\
Solar & 720 & 0.3536 & 0.2271 & 0.3002 & 0.3079 & 0.2561 \\
Weather & 96 & 0.1999 & 0.1718 & 0.1727 & 0.1697 & 0.1827 \\
Weather & 192 & 0.2397 & 0.2189 & 0.2204 & 0.2176 & 0.2277 \\
Weather & 336 & 0.2904 & 0.2679 & 0.2753 & 0.2774 & 0.2819 \\
Weather & 720 & 0.3652 & 0.3275 & 0.3507 & 0.3539 & 0.3595 \\
\bottomrule
\end{tabular}

\caption{Phase-1 forecasting benchmark: test MSE. Hybrid-0 is included only to document the architecture-screening stage; the current paper does not claim it as a competitive final forecaster.}
\label{tab:app-baseline-mse}
\end{table}
\begin{table}[!ht]
\centering
\small
\begin{tabular}{lrrrrrr}
\toprule
Dataset & H & DLinear & Hybrid-0 & PatchTST & TimesNet & iTransformer \\
\midrule
ETTh1 & 96 & 0.4030 & 0.4346 & 0.4033 & 0.4132 & 0.4095 \\
ETTh1 & 192 & 0.4507 & 0.4830 & 0.4333 & 0.4497 & 0.4294 \\
ETTh1 & 336 & 0.4784 & 0.5731 & 0.4604 & 0.4978 & 0.4520 \\
ETTh1 & 720 & 0.5283 & 0.6451 & 0.4900 & 0.5078 & 0.4788 \\
ETTm1 & 96 & 0.3704 & 0.4092 & 0.3694 & 0.3681 & 0.3742 \\
ETTm1 & 192 & 0.4011 & 0.4372 & 0.3919 & 0.4044 & 0.3910 \\
ETTm1 & 336 & 0.4192 & 0.4872 & 0.4133 & 0.4258 & 0.4204 \\
ETTm1 & 720 & 0.4713 & 0.5086 & 0.4461 & 0.4786 & 0.4638 \\
Electricity & 96 & 0.2780 & 0.2861 & 0.3021 & 0.2843 & 0.2437 \\
Electricity & 192 & 0.2809 & 0.2972 & 0.3102 & 0.2976 & 0.2558 \\
Electricity & 336 & 0.2963 & 0.3142 & 0.3237 & 0.3122 & 0.2723 \\
Electricity & 720 & 0.3332 & 0.3483 & 0.3542 & 0.3372 & 0.3031 \\
Solar & 96 & 0.3724 & 0.2729 & 0.3194 & 0.2491 & 0.2372 \\
Solar & 192 & 0.3926 & 0.2857 & 0.3359 & 0.2965 & 0.2657 \\
Solar & 336 & 0.4133 & 0.3037 & 0.3423 & 0.2943 & 0.2767 \\
Solar & 720 & 0.4130 & 0.2858 & 0.3513 & 0.3191 & 0.2796 \\
Weather & 96 & 0.2662 & 0.2372 & 0.2159 & 0.2183 & 0.2246 \\
Weather & 192 & 0.2999 & 0.2783 & 0.2571 & 0.2606 & 0.2621 \\
Weather & 336 & 0.3447 & 0.3191 & 0.2959 & 0.3003 & 0.3009 \\
Weather & 720 & 0.4065 & 0.3500 & 0.3447 & 0.3508 & 0.3524 \\
\bottomrule
\end{tabular}

\caption{Phase-1 forecasting benchmark: test MAE.}
\label{tab:app-baseline-mae}
\end{table}
\begin{table}[!ht]
\centering
\small
\setlength{\tabcolsep}{3.5pt}
\begin{tabular}{lrrrrrr}
\toprule
model & mean\_mse\_rank & median\_mse\_rank & mean\_mae\_rank & mse\_best & mse\_top2 & mae\_best \\
\midrule
iTransformer & 2.45 & 2.00 & 1.85 & 5 & 12 & 12 \\
PatchTST & 2.70 & 2.50 & 2.70 & 6 & 10 & 6 \\
TimesNet & 2.95 & 3.00 & 2.95 & 3 & 5 & 1 \\
Hybrid-0 & 3.25 & 4.00 & 3.95 & 6 & 8 & 0 \\
DLinear & 3.65 & 3.50 & 3.55 & 0 & 5 & 1 \\
\bottomrule
\end{tabular}

\caption{Aggregate rank summary over 20 conditions. iTransformer has the best mean MSE rank; TimesNet is the architecture-distinct confirmation backbone.}
\label{tab:app-baseline-rank}
\end{table}

The primary strong-backbone experiments use iTransformer because it obtains the best mean MSE rank (2.45) and best mean MAE rank (1.85) among the five displayed systems. TimesNet is retained as the confirmatory architecture because it differs substantially in inductive bias while remaining competitive enough to make a meaningful model-dependence test.

\section{Explicit Hard Sparse iTransformer Intervention}
\label{app:hardmask}
The main-paper transfer experiment directly restricts the final cross-channel interaction layer. DenseFineTune leaves channel interaction unrestricted, SharedSparse uses a horizon-invariant Borda consensus topology, and HorizonAdaptiveSparse uses the predictive-utility ranking for the current prediction length. All variants start from the same Phase-1 iTransformer checkpoint. We evaluate all 20 conditions from Electricity, Weather, Solar, ETTh1, and ETTm1 for which the matched Phase-1 checkpoint suite is available.

\begin{table}[!ht]
\centering
\small
\begin{tabular}{lrrrr}
\toprule
Dataset & $H=96$ & $H=192$ & $H=336$ & $H=720$\\
\midrule
ETTh1       & -0.2262 & +0.2202 & -0.0384 & -0.0125\\
ETTm1       & +0.0577 & -0.1938 & +3.3341 & +0.1088\\
Electricity & -0.6678 & -0.4854 & +0.7314 & -0.7349\\
Solar       & +2.2269 & +0.1286 & +0.3790 & +0.1349\\
Weather     & -1.0549 & +0.2322 & +0.6420 & -0.6224\\
\bottomrule
\end{tabular}
\caption{HorizonAdaptiveSparse-versus-SharedSparse full-horizon-MSE gain (\%) across the complete 20-condition matched iTransformer checkpoint suite. Positive values favor the horizon-specific topology.}
\label{tab:app-hardmask-all20}
\end{table}

HorizonAdaptiveSparse wins 11/20 conditions under full-horizon MSE, with mean gain $+0.208\%$ and median $+0.083\%$. The condition-wise gain is almost uncorrelated with the controlled Ridge Adaptive-versus-Shared gain ($\rho=0.057$), and the gain signs match in only 10/20 conditions. Endpoint MSE gives the same 11/20 win count but mean gain $-0.120\%$, median $+0.070\%$, correlation $\rho=-0.074$, and sign agreement 8/20. Thus, expanding the original diagnostic screen to the full matched checkpoint suite does not reveal a reliable transfer relation.

The strongest controlled cases remain illustrative. ETTh1 $H=336$ and $720$ have controlled gains $+15.1233\%$ and $+17.2641\%$, yet their full-horizon neural gains are $-0.0384\%$ and $-0.0125\%$. ETTm1 $H=720$ changes from $+9.4491\%$ controlled gain to only $+0.1088\%$ under the hard mask. Conversely, some conditions with negative controlled gain become positive under iTransformer (e.g., Solar $H=96$). These reversals are why the paper emphasizes transfer alignment rather than whether the mean neural gain is exactly zero.

\section{Robustness Across Alternative Horizon-Conditioning Mechanisms}
\label{app:alternative-routing}
\label{app:alternative}

The hard mask is not the first conditioning mechanism we evaluated. To rule out the explanation that the transfer failure is peculiar to one intervention, we tested multiple softer ways of injecting horizon-specific structure. We report these as robustness checks, not as a post-hoc architecture search.

\begin{table}[!ht]
\centering
\small
\begin{tabular}{lllr}
\toprule
Mechanism & Compared against & Wins & Mean advantage (pp) \\
\midrule
Horizon gate & Global gate & 3/4 & 0.0443 \\
Horizon query & Horizon-invariant & 0/4 & -0.4423 \\
Grouped query & Shared query & 1/4 & -0.2234 \\
Predictive bias & Shared bias & 0/4 & -0.0229 \\
Hard sparse mask (8-cond. screen) & Shared sparse & 4/8 & 0.0240 \\
\bottomrule
\end{tabular}

\caption{Alternative horizon-conditioning mechanisms. ``Mean advantage'' is the horizon-specific variant minus its matched shared/invariant control, in percentage points of MSE gain.}
\label{tab:app-alt-summary}
\end{table}
\begin{table}[!ht]
\centering
\small
\begin{tabular}{lrrrrr}
\toprule
Dataset & H & Horizon gate & Horizon query & Grouped query & Predictive bias \\
\midrule
ETTh1 & 336 & 0.1363 & -0.3545 & 0.0154 & -0.0192 \\
ETTm1 & 336 & 0.0914 & -1.3197 & -0.8756 & -0.0279 \\
Electricity & 192 & 0.0174 & -0.0109 & -0.0332 & -0.0234 \\
Solar & 720 & -0.0678 & -0.0842 & 0.0000 & -0.0210 \\
\bottomrule
\end{tabular}

\caption{Per-condition advantage (percentage points) of four horizon-specific mechanisms over their matched shared/invariant controls. No soft mechanism produces a consistent transfer effect.}
\label{tab:app-alt-conditions}
\end{table}

The horizon-gated mixture gives a small positive mean advantage over a global gate (0.044 percentage points, 3/4 conditions), but remains worse than previously tested non-gated alternatives on several conditions. Sparse HorizonQuery is worse than its horizon-invariant control in all four conditions. The grouped query adapter wins only 1/4, and horizon-grouped predictive attention bias is worse than shared bias in all four. These results motivated the final hard-mask experiment: if the model could simply ignore weak soft conditioning, a mandatory structural intervention should expose the benefit. It did not.

\section{Grouped Channel-Permutation Analysis}
\label{app:grouped-permutation}

We perturb channel histories using cyclic donor mappings. For a grouped intervention, all selected source channels use the same donor sample, preserving source--source coherence while breaking their alignment with the current target context. This is a model intervention, not a causal intervention on the data-generating process.

\begin{figure}[!t]
\centering
\includegraphics[width=0.78\textwidth]{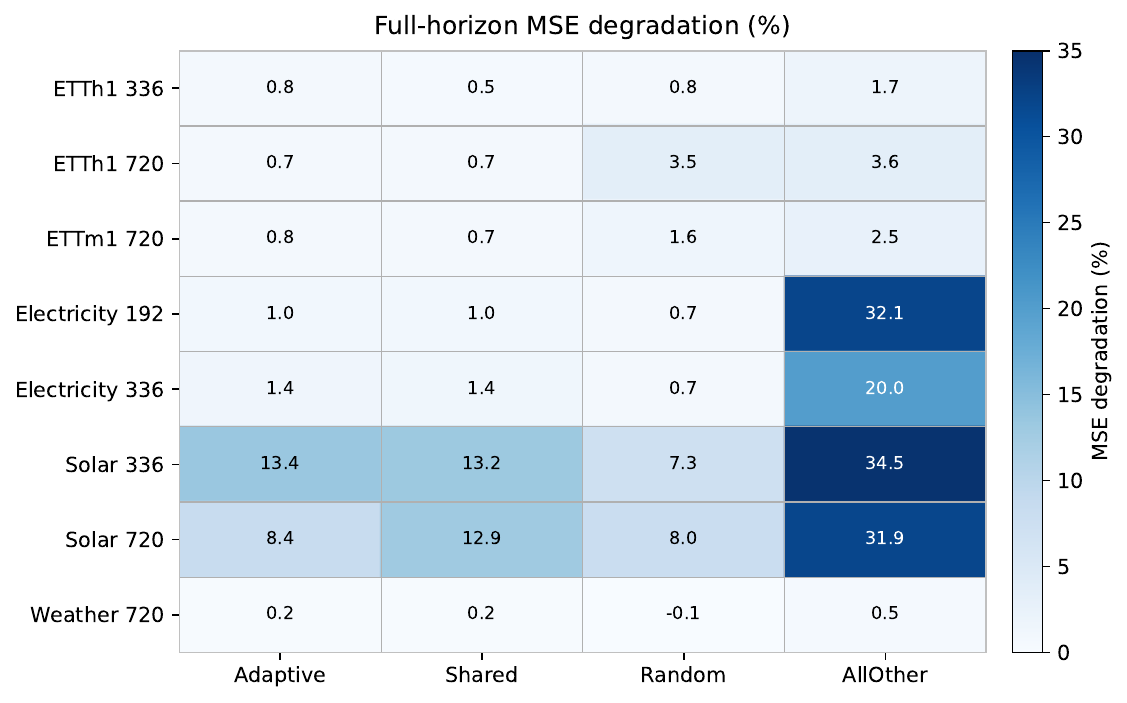}
\caption{\textbf{Full-horizon MSE degradation under grouped source interventions.} Cross-channel information matters in all eight tested conditions, but the relative ordering of Adaptive, Shared, and matched Random subsets is not consistent.}
\label{fig:app-group-perm}
\end{figure}
\begin{table}[!ht]
\centering
\small
\begin{tabular}{lrrrrr}
\toprule
Dataset & H & Adaptive & Shared & Random & AllOther \\
\midrule
ETTh1 & 336 & 0.755 & 0.529 & 0.766 & 1.697 \\
ETTh1 & 720 & 0.698 & 0.715 & 3.470 & 3.552 \\
ETTm1 & 720 & 0.811 & 0.654 & 1.569 & 2.540 \\
Electricity & 192 & 0.998 & 0.982 & 0.686 & 32.100 \\
Electricity & 336 & 1.421 & 1.358 & 0.690 & 20.025 \\
Solar & 336 & 13.438 & 13.221 & 7.303 & 34.490 \\
Solar & 720 & 8.422 & 12.924 & 7.978 & 31.863 \\
Weather & 720 & 0.171 & 0.156 & -0.094 & 0.466 \\
\bottomrule
\end{tabular}

\caption{Grouped source-history permutation: full-horizon MSE degradation (\%). AllOther perturbation preserves the target history but replaces every non-target channel using a common donor mapping.}
\label{tab:app-perm-full}
\end{table}
\begin{table}[!ht]
\centering
\small
\begin{tabular}{lrrrrrr}
\toprule
Dataset & H & Adaptive & AllOther & Random & Shared & TargetSelf \\
\midrule
ETTh1 & 336 & 0.697 & 1.869 & 1.067 & 0.379 & 92.756 \\
ETTh1 & 720 & 0.584 & 4.008 & 4.015 & 0.544 & 82.066 \\
ETTm1 & 720 & -0.469 & 1.671 & 1.069 & 0.367 & 50.837 \\
Electricity & 192 & 1.394 & 53.315 & 1.106 & 1.398 & 431.977 \\
Electricity & 336 & 1.964 & 7.750 & 1.054 & 1.955 & 305.716 \\
Solar & 336 & 5.916 & 17.538 & 4.185 & 6.357 & 14.139 \\
Solar & 720 & 3.563 & 19.412 & 4.097 & 5.941 & 12.350 \\
Weather & 720 & -0.164 & -0.227 & -0.460 & -0.093 & 1.048 \\
\bottomrule
\end{tabular}

\caption{Grouped source-history permutation: endpoint MSE degradation (\%). TargetSelf is a sanity check showing the much larger dependence on the target's own history.}
\label{tab:app-perm-endpoint}
\end{table}

AllOtherChannels increases full-horizon MSE in 8/8 conditions, with mean degradation $+15.842\%$. The mean degradations for AdaptiveTopK, SharedTopK, and MatchedRandomK are $+3.339\%$, $+3.817\%$, and $+2.796\%$, respectively. The result rules out the trivial explanation that iTransformer does not use non-target channels. At the same time, Adaptive does not dominate either Shared or Random, reinforcing the source-identification mismatch.

Electricity illustrates distributed or redundant dependence: at $H=192$, AllOther endpoint degradation is 53.315\%, while individual candidate-source effects are small on average. Solar is more concentrated, with both large AllOther degradation and much larger individual-source importance. The concentration pattern therefore varies by dataset.

\section{Individual Dependency, Utility, and Neural Reliance}
\label{app:individual-api}

The individual-source analysis evaluates 737 candidate pairs over 55 target channels in eight representative conditions. Three source-level signals are compared with endpoint functional importance: controlled predictive utility $P$, raw-space absolute dependency $A_{\rm raw}$, and final-encoder latent absolute cosine dependency $A_{\rm latent}$.

\begin{figure}[!t]
\centering
\includegraphics[width=0.62\textwidth]{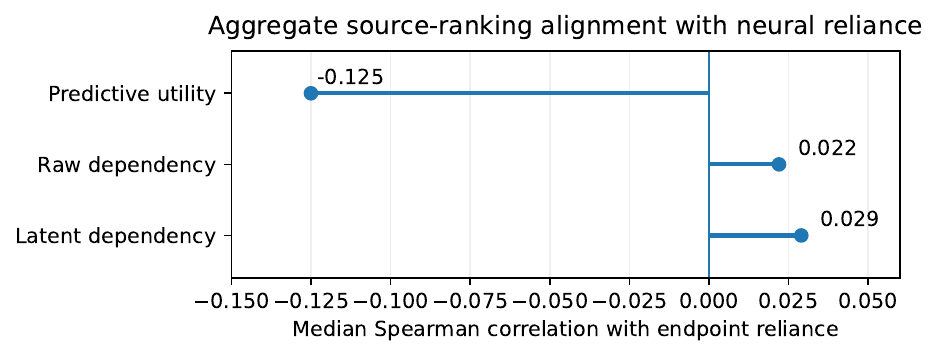}
\caption{\textbf{Global source-ranking alignment with neural reliance.} None of the three static source scores shows strong rank agreement with endpoint functional reliance. Raw and latent dependency provide somewhat better top-heavy overlap than predictive utility, but median Spearman correlations remain near zero.}
\label{fig:app-static-align}
\end{figure}
\begin{table}[!ht]
\centering
\small
\begin{tabular}{lrrrr}
\toprule
Dataset/$H$ & Targets & $\rho(P,I)$ & $\rho(A_{\rm raw},I)$ & $\rho(A_{\rm latent},I)$ \\
\midrule
ETTh1/336 & 7 & -0.029 & -0.143 & -0.257 \\
ETTh1/720 & 7 & -0.371 & -0.086 & -0.086 \\
ETTm1/720 & 7 & -0.314 & 0.429 & 0.029 \\
Electricity/192 & 6 & 0.153 & -0.134 & 0.064 \\
Electricity/336 & 6 & -0.203 & 0.088 & -0.039 \\
Solar/336 & 6 & -0.068 & 0.083 & 0.024 \\
Solar/720 & 6 & -0.146 & 0.051 & 0.042 \\
Weather/720 & 10 & 0.036 & 0.081 & 0.277 \\
\bottomrule
\end{tabular}
\vspace{2mm}

\begin{tabular}{lrrrrr}
\toprule
Dataset/$H$ & $J(P,I)$ & $J(A_{\rm raw},I)$ & $J(A_{\rm latent},I)$ & Mean $I$ & Seed std. \\
\midrule
ETTh1/336 & 0.810 & 0.762 & 0.762 & 0.210 & 0.258 \\
ETTh1/720 & 0.667 & 0.667 & 0.762 & 0.368 & 0.504 \\
ETTm1/720 & 0.667 & 0.762 & 0.810 & 0.857 & 0.883 \\
Electricity/192 & 0.210 & 0.139 & 0.181 & 0.029 & 0.325 \\
Electricity/336 & 0.093 & 0.222 & 0.217 & 0.127 & 0.600 \\
Solar/336 & 0.111 & 0.139 & 0.144 & 4.612 & 3.066 \\
Solar/720 & 0.093 & 0.316 & 0.286 & 8.783 & 2.887 \\
Weather/720 & 0.119 & 0.269 & 0.335 & 0.048 & 0.453 \\
\bottomrule
\end{tabular}
\caption{Condition-level alignment between static source scores and endpoint functional reliance. Top: rank correlations. Bottom: Top-$K$ overlaps and importance magnitude.}
\label{tab:app-api-summary}
\end{table}

Aggregating the eight conditions gives median Spearman $-0.125$ for predictive utility, $0.022$ for raw dependency, and $0.029$ for latent dependency. Mean Top-$K$ Jaccard is 0.350, 0.417, and 0.448, respectively; mean recall is 0.447, 0.531, and 0.560. Thus, top-heavy set overlap can be moderate even when the complete rankings are weakly aligned.

\subsection{Cross-prediction-length stability}
\begin{table}[!ht]
\centering
\small
\begin{tabular}{lrrrrrr}
\toprule
Dataset & $H_1$ & $H_2$ & Targets & $\rho_P$ & $\rho_I$ & $\rho_{\rm latent}$ \\
\midrule
ETTh1 & 336 & 720 & 7 & 0.714 & 0.771 & 0.771 \\
Electricity & 192 & 336 & 6 & 0.699 & -0.135 & 0.993 \\
Solar & 336 & 720 & 6 & 0.417 & -0.192 & 0.936 \\
\bottomrule
\end{tabular}
\vspace{2mm}

\begin{tabular}{lrrrrr}
\toprule
Dataset & $H_1$ & $H_2$ & $J_P$ & $J_I$ & $J_{\rm latent}$ \\
\midrule
ETTh1 & 336 & 720 & 0.905 & 0.857 & 1.000 \\
Electricity & 192 & 336 & 0.339 & 0.120 & 0.833 \\
Solar & 336 & 720 & 0.116 & 0.097 & 0.643 \\
\bottomrule
\end{tabular}
\caption{Cross-prediction-length stability. Top: median rank correlations. Bottom: mean Top-$K$ Jaccard. Checkpoints are trained separately for each prediction length, so these comparisons are descriptive.}
\label{tab:app-api-cross}
\end{table}

The mean cross-horizon Jaccard values across the three matched dataset pairs are approximately $J_P=0.453$, $J_I=0.358$, and $J_{\rm latent}=0.825$. We do \emph{not} interpret this as a causal effect of horizon because each prediction length uses a separately trained checkpoint. The result instead shows that latent representation dependency can remain highly stable even when functional source reliance changes substantially.

\subsection{Same-checkpoint future-offset reliance}
\label{app:same-checkpoint}
The cross-prediction-length comparison above uses separately trained checkpoints. To remove that confound, we fix the $H=720$ iTransformer checkpoint for each of the five general datasets and evaluate source reliance at offsets $h\in\{96,192,336,720\}$. The target--source candidate pools, sampled test windows, and perturbation mappings are held fixed across offsets. This yields 36 target--condition instances for each offset comparison.

Table~
ef{tab:same-checkpoint-main} in the main paper reports the complete offset-pair stability and offset-specific utility--reliance alignment summaries.

Functional reliance is therefore neither fully stable nor arbitrary across offsets: pairwise median correlations range from 0.148 to 0.482. More importantly for the paper's main claim, $P$--EPI alignment remains negative at every offset, as summarized in Table~\ref{tab:same-checkpoint-main}. The utility--reliance mismatch is thus visible inside a single fixed forecaster and cannot be attributed solely to independently optimized prediction-length checkpoints.

\section{Utility Definition and Candidate-Pool Robustness}
\label{app:utility-robustness}

The initial source-selection discovery uses endpoint MSE. A natural concern is that poor $P$--$I$ alignment might therefore be caused by comparing endpoint utility with full-horizon neural importance, or by evaluating only a utility/dependency-enriched candidate pool. The final robustness study recomputes utility for endpoint MSE, full-horizon MSE, and full-horizon MAE and augments the candidate set with deterministic random sources.

\begin{table}[!ht]
\centering
\small
\begin{tabular}{llrrr}
\toprule
Pool & Metric & Targets & Median $\rho(P,\mathrm{EPI})$ & Mean $\rho(P,\mathrm{EPI})$ \\
\midrule
BaseOnly & endpoint MSE & 19 & -0.011 & -0.098 \\
BaseOnly & full MAE & 19 & -0.373 & -0.246 \\
BaseOnly & full MSE & 19 & -0.371 & -0.303 \\
Extended & endpoint MSE & 19 & -0.164 & -0.109 \\
Extended & full MAE & 19 & -0.320 & -0.283 \\
Extended & full MSE & 19 & -0.354 & -0.325 \\
\bottomrule
\end{tabular}
\vspace{2mm}

\begin{tabular}{llrr}
\toprule
Pool & Metric & Median $\rho(I_{\rm old},\mathrm{EPI})$ & Mean $J(P,\mathrm{EPI})$ \\
\midrule
BaseOnly & endpoint MSE & 0.543 & 0.338 \\
BaseOnly & full MAE & 0.639 & 0.306 \\
BaseOnly & full MSE & 0.702 & 0.300 \\
Extended & endpoint MSE & 0.500 & 0.281 \\
Extended & full MAE & 0.623 & 0.301 \\
Extended & full MSE & 0.663 & 0.282 \\
\bottomrule
\end{tabular}
\caption{Predictive-utility alignment after stabilizing iTransformer importance with 16-donor EPI. Top: predictive-utility correlations. Bottom: agreement with the earlier importance estimate and Top-$K$ overlap.}
\label{tab:app-epi-pool}
\end{table}

Under the final 16-donor EPI, the Extended-pool median correlation between predictive utility and iTransformer reliance is $-0.164$ for endpoint MSE, $-0.354$ for full MSE, and $-0.320$ for full MAE. BaseOnly gives $-0.011$, $-0.371$, and $-0.373$, respectively. The mismatch is therefore not rescued by matching the forecasting objective or by broadening the candidate pool.

The earlier five-donor importance estimate and the final 16-donor EPI have median correlations of 0.500 for endpoint MSE, 0.663 for full MSE, and 0.623 for full MAE. This is another reason to use full-horizon MSE as the primary functional-reliance measure.

\section{Expected Permutation Importance Reliability}
\label{app:epi-reliability}

For every evaluated source, EPI averages the same 16 nonzero donor mappings. We separately measure: (i) donor split-half reliability by dividing the 16 donors into 8/8 halves, (ii) random-window split-half reliability over 50 repetitions, and (iii) temporal-block stability over four contiguous test blocks. Reliability is measured as target-wise Spearman agreement of source rankings and summarized by the median across targets.

\begin{figure}[!t]
\centering
\includegraphics[width=0.95\textwidth]{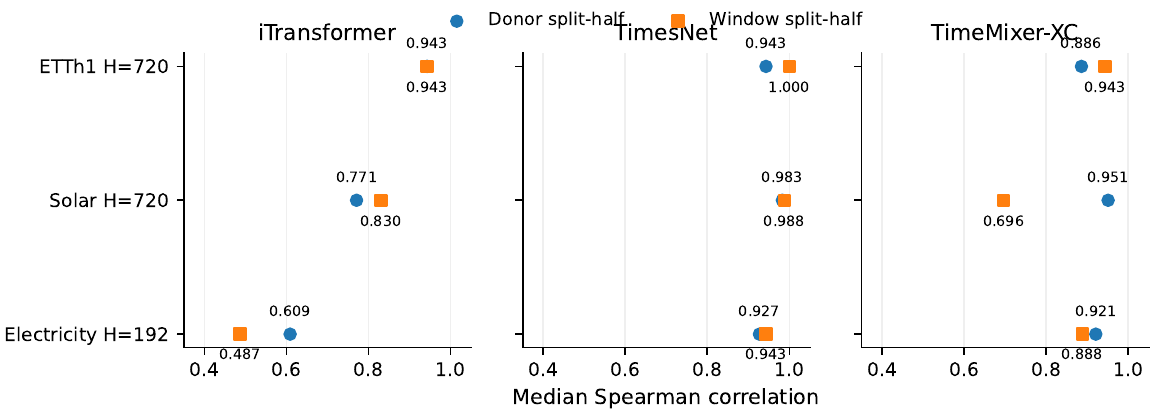}
\caption{\textbf{Full-MSE EPI split-half reliability.} Donor and random-window split-half correlations are shown for iTransformer, TimesNet, and cross-channel TimeMixer on the same three confirmatory conditions. All three are reproducible under donor resampling; window stability is lower for iTransformer on Electricity and TimeMixer on Solar.}
\label{fig:app-epi-rel}
\end{figure}
\begin{table}[!ht]
\centering
\small
\begin{tabular}{lrlrrrr}
\toprule
Dataset & H & Metric & Targets & Donor split & Window split & Temporal blocks \\
\midrule
ETTh1 & 720 & endpoint\_mse & 7 & 0.943 & 0.943 & 0.362 \\
ETTh1 & 720 & full\_mae & 7 & 0.943 & 0.943 & 0.533 \\
ETTh1 & 720 & full\_mse & 7 & 0.943 & 0.943 & 0.533 \\
Electricity & 192 & endpoint\_mse & 6 & 0.409 & 0.184 & -0.068 \\
Electricity & 192 & full\_mae & 6 & 0.544 & 0.449 & 0.024 \\
Electricity & 192 & full\_mse & 6 & 0.609 & 0.487 & 0.007 \\
Solar & 720 & endpoint\_mse & 6 & 0.348 & 0.058 & 0.030 \\
Solar & 720 & full\_mae & 6 & 0.882 & 0.908 & 0.755 \\
Solar & 720 & full\_mse & 6 & 0.771 & 0.830 & 0.658 \\
\bottomrule
\end{tabular}

\caption{iTransformer EPI reliability. Values are median target-wise Spearman correlations under donor split-half, random-window split-half, and temporal-block comparisons.}
\label{tab:app-it-reliability}
\end{table}

Endpoint importance is clearly less reliable in the high-dimensional conditions: donor/window reliability is 0.409/0.184 on Electricity and 0.348/0.058 on Solar. Full-MSE reliability improves to 0.609/0.487 and 0.771/0.830, respectively; ETTh1 is 0.943/0.943. Full MAE gives a similar pattern. We therefore treat endpoint results as secondary and use full-horizon MSE for the primary model-reliance claims.

Temporal stability is not universal. It is close to zero for Electricity under iTransformer and TimesNet, high for Solar under TimesNet, but only $0.061$ for TimeMixer on Solar; ETTh1 is moderate for all three. We retain temporal-block stability as a secondary domain-dependent observation rather than a central conclusion.

\section{Cross-Backbone Functional-Reliance Confirmation}
\label{app:timesnet}

TimesNet is evaluated without retraining or outcome-dependent tuning. The Phase-1 best checkpoint is loaded, and the exact iTransformer EPI target--source pairs, 512 test origins, and 16 donor mappings are reused. TimeMixer is trained separately with explicit cross-channel mixing under the fixed configuration documented in Appendix~\ref{app:protocol}, then evaluated on the same pairs, origins, and mappings. This matched intervention protocol isolates forecaster dependence more directly than comparisons with the compact Ridge/MLP probes.

\begin{figure}[!t]
\centering
\includegraphics[width=0.62\textwidth]{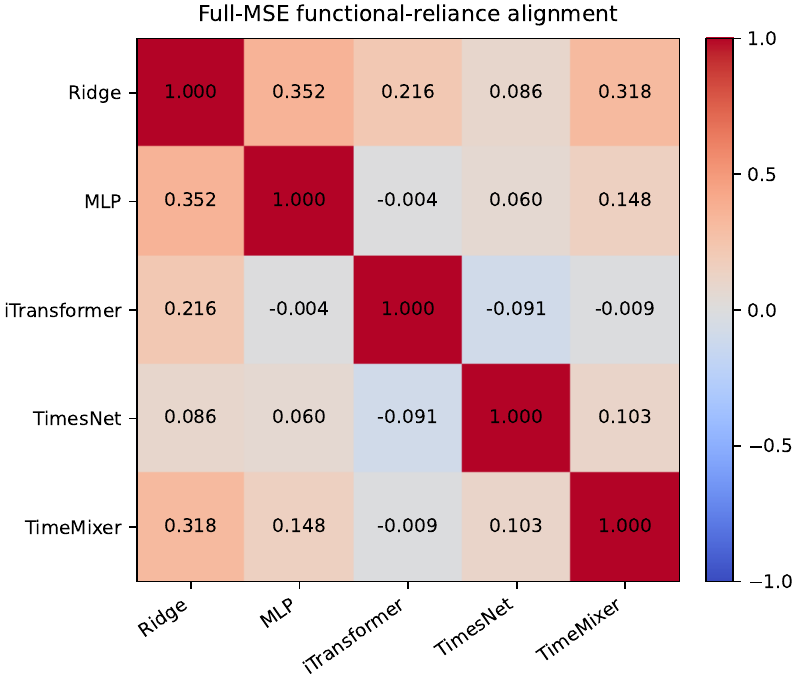}
\caption{\textbf{Median full-MSE functional-reliance alignment across forecasters.} The three neural pairs---iTransformer--TimesNet, iTransformer--TimeMixer, and TimesNet--TimeMixer---have median target-wise Spearman correlations $-0.091$, $-0.009$, and $0.103$, respectively.}
\label{fig:app-forecaster-corr}
\end{figure}

\begin{table}[!ht]
\centering
\small
\begin{tabular}{llrrrr}
\toprule
Forecaster A & Forecaster B & Targets & Median $\rho$ & Mean $\rho$ & Top-5 Jaccard \\
\midrule
MLP & TimesNet & 19 & 0.060 & 0.136 & 0.368 \\
MLP & iTransformer & 19 & -0.004 & 0.082 & 0.431 \\
MLP & TimeMixer & 19 & 0.148 & 0.074 & 0.310 \\
Ridge & MLP & 19 & 0.352 & 0.330 & 0.457 \\
Ridge & TimesNet & 19 & 0.086 & 0.152 & 0.354 \\
Ridge & iTransformer & 19 & 0.216 & 0.084 & 0.352 \\
Ridge & TimeMixer & 19 & 0.318 & 0.291 & 0.374 \\
iTransformer & TimesNet & 19 & -0.091 & 0.045 & 0.348 \\
iTransformer & TimeMixer & 19 & -0.009 & -0.032 & 0.301 \\
TimesNet & TimeMixer & 19 & 0.103 & 0.100 & 0.357 \\
\bottomrule
\end{tabular}
\caption{Full-horizon-MSE forecaster-to-forecaster functional-reliance alignment under the final matched candidate set.}
\label{tab:app-forecaster-pairs}
\end{table}

\begin{table}[!ht]
\centering
\small
\begin{tabular}{lrrrr}
\toprule
Dataset & $H$ & Donor split & Window split & Temporal blocks \\
\midrule
Electricity & 192 & 0.927 & 0.943 & 0.033 \\
Solar & 720 & 0.983 & 0.988 & 0.945 \\
ETTh1 & 720 & 0.943 & 1.000 & 0.333 \\
\bottomrule
\end{tabular}
\caption{TimesNet EPI reliability under the same test origins, target--source pairs, and donor mappings used for iTransformer.}
\label{tab:app-tn-reliability}
\end{table}

\begin{table}[!ht]
\centering
\small
\setlength{\tabcolsep}{4pt}
\begin{tabular}{lrrrrrr}
\toprule
Dataset & $H$ & Test MSE & Test MAE & Grouped med. & Donor & Window \\
\midrule
Electricity & 192 & 0.1658 & 0.2665 & 147.599 & 0.921 & 0.888 \\
Solar & 720 & 0.2796 & 0.3062 & 1015.435 & 0.951 & 0.696 \\
ETTh1 & 720 & 0.5658 & 0.5357 & 27.426 & 0.886 & 0.943 \\
\bottomrule
\end{tabular}
\caption{Cross-channel TimeMixer confirmation. ``Grouped med.'' is the median full-MSE increase (\%) when all non-target histories are replaced together; all 19 evaluated targets have positive grouped effects. The large grouped magnitudes are used only to verify cross-channel sensitivity, not as causal effect sizes. Temporal-block rank correlations are 0.230, 0.061, and 0.429 for Electricity, Solar, and ETTh1.}
\label{tab:app-tm-confirm}
\end{table}

Under full-horizon MSE,
\[
\rho(P,I^{\rm TimesNet})=0.029,\qquad
\rho(P,I^{\rm TimeMixer})=0.053.
\]
The neural pairwise median correlations are
\[
\begin{aligned}
\rho(I^{\rm iTransformer},I^{\rm TimesNet})&=-0.091, &
\rho(I^{\rm iTransformer},I^{\rm TimeMixer})&=-0.009,\\
\rho(I^{\rm TimesNet},I^{\rm TimeMixer})&=0.103.
\end{aligned}
\]
Thus, adding a third architecture does not reveal a common functional source ranking. TimeMixer donor/window split-half reliability is 0.921/0.888 on Electricity, 0.951/0.696 on Solar, and 0.886/0.943 on ETTh1; TimesNet remains 0.927/0.943, 0.983/0.988, and 0.943/1.000. The cross-neural disagreement therefore persists despite reproducible within-model EPI estimates.

Comparisons involving Ridge and MLP remain broader forecaster-level diagnostics because their compact input representations differ from the neural models. The matched iTransformer--TimesNet--TimeMixer comparison is the cleanest evidence for architecture conditioning: it holds the target--source pairs, test windows, and donor mappings fixed while changing the trained forecasting mechanism.

\section{Locked Bounded-Support Protocol}
\label{app:bounded-support}
The constructive study uses an official channel-independent PatchTST as the frozen base. For its patch latents $\mathbf z\in\mathbb R^{B\times C\times P\times D}$, the adapter performs channel self-attention independently at each patch, applies a residual feed-forward block, and projects the change $\mathbf z^{\rm mix}-\mathbf z$ to the forecast length. A $0.5\tanh(\cdot)$ output bound, a sigmoid gate for each channel and 24-step forecast chunk, and the base model's scale restore the residual $\Delta_\theta(\mathbf X)$ in observation units. Setting every gate or $\alpha$ to zero recovers the frozen base exactly. The adapter is trained on the fit prefix while the base is frozen; a disjoint chronological calibration tail supplies the coefficients below.

Let $\mathbf E=\widehat{\mathbf Y}^{\rm base}-\mathbf Y$ and $\mathbf D=\Delta_\theta(\mathbf X)$ on that calibration tail. Global calibration uses
\begin{equation}
\alpha_g=\operatorname{clip}_{[0,1]}
\left(-\frac{\langle\mathbf E,\mathbf D\rangle}
{\lVert\mathbf D\rVert_2^2}\right).
\end{equation}
For channel $c$, the same estimator gives $\widetilde\alpha_c$. With correction energy $q_c=\sum_{n,h}D_{nhc}^2$, median energy $q_{\rm med}$, and $w_c=q_c/(q_c+q_{\rm med})$, channel-shrunk calibration uses $\alpha_c=w_c\widetilde\alpha_c+(1-w_c)\alpha_g$. For cross-fitting, the calibration tail is split into two chronological halves. Each half estimates an $\alpha_g$ that is retained only if it lowers MSE on the opposite half; accepted coefficients are averaged, with $\alpha=0$ if neither is accepted. This makes the fallback rule explicit without test-dependent selection.

The evaluation grid contains four datasets (ETTh2, ETTm1, Exchange, and Weather), four prediction lengths (96, 192, 336, and 720), and five seeds. The test split is evaluated once after fixing all policies and the primary comparison. Aggregate uncertainty uses hierarchical resampling over datasets, horizons, and seeds. The $0.5\%$ statement is a descriptive diagnostic over five-seed dataset--horizon cell means, not a formal non-inferiority test. The largest observed seed-level regression is $1.261\%$, so the result supports positive aggregate transfer rather than a per-seed or formal non-degradation guarantee. Complete per-seed outputs are retained in the frozen evaluation record.

\section{Interpretation and Scope}
\label{app:scope}

\paragraph{What the experiments do not establish.}
We do not claim that every horizon-adaptive channel model is ineffective, that cross-channel information is unimportant, or that predictive relevance is intrinsically model-dependent. The controlled utility $P$ remains one transparent Ridge-based operational probe, not a model-free population quantity; the MLP experiment tests the transfer of its selected source sets rather than independently redefining $P$. The high-dimensional robustness study further shows that its aggregate gains can depend on the evaluated target population and admissible source pool. EPI is a functional intervention on a fixed trained model and does not identify causal relations among physical variables. The same-checkpoint offset study removes one optimization confound, but it still measures model sensitivity rather than causal data-generating relations.

\paragraph{Scope of architecture evaluation.}
The conclusion is tested through complementary controls rather than open-ended architecture search: a nonlinear controlled predictor, the complete 20-condition matched hard-mask suite, a same-checkpoint offset analysis, several soft conditioning mechanisms, high-dimensional target/candidate sensitivity checks, and matched EPI confirmations on two additional neural architectures. We therefore stop at three neural forecasters while keeping the architecture-level claim explicitly limited to the evaluated models.

\paragraph{Reproducibility map.}
The public reproducibility repository at \url{https://github.com/dearyonghoon/rethinking-cross-channel} provides executable notebooks for cross-domain drift, corrected controlled selection, the nonlinear MLP confirmation, high-dimensional target-subset/candidate-cap sensitivity, the 20-condition hard sparse iTransformer intervention, grouped and individual source interventions, same-checkpoint offset EPI, EPI reliability, TimesNet and TimeMixer confirmations, and the Phase-1 baseline benchmark, together with frozen result summaries where available.

\end{document}